\documentclass[letterpaper,twocolumn,10pt]{article}
\usepackage{usenix-2020-09}

\usepackage{epsfig,amsmath,amsfonts,epsfig,multirow,makecell,caption,soul,csquotes,color,wrapfig,subcaption,mathtools,bm,spverbatim,booktabs,tcolorbox,diagbox,todonotes,amsthm}
\usepackage[e]{esvect}

\usepackage{accents}

\usepackage{caption}
\makeatletter
\newif\if@restonecol
\makeatother

\usepackage[boxed, ruled, vlined, linesnumbered]{algorithm2e}
\SetKwRepeat{Do}{do}{while}

\usepackage[first=0,last=9]{lcg}
\usepackage{colortbl}
\definecolor{Gray}{gray}{0.8}
\colorlet{Red}{red!10!white}
\colorlet{Blue}{blue!10!white}

\newenvironment{changemargin}[2]{\begin{list}{}{
	\setlength{\topsep}{0pt}\setlength{\leftmargin}{0pt}
	\setlength{\rightmargin}{0pt}
	\setlength{\listparindent}{\parindent}
	\setlength{\itemindent}{\parindent}
	\setlength{\parsep}{0pt plus 1pt}
	\addtolength{\leftmargin}{#1}\addtolength{\rightmargin}{#2}
	}\item}
	{\end{list}}

\newcommand{\ssup}[2]{{#1}^{\scaleobj{0.8}{#2}}}
\newcommand{\ssub}[2]{{#1}_{\scaleobj{0.8}{#2}}}
\newcommand{\sboth}[3]{{#1}_{\scaleobj{0.8}{#2}}^{\scaleobj{0.8}{#3}}}

\usepackage[first=0,last=9]{lcg}
\usepackage{colortbl}
\definecolor{Gray}{gray}{0.8}
\definecolor{BrickRed}{RGB}{203,65,84}

\newtheorem{theorem}{Theorem}

\usepackage{hyperref}

\newcommand{\msec}[1]{\S\,\ref{#1}}
\newcommand{\mref}[1]{\,\ref{#1}}
\newcommand{\meq}[1]{Eqn\,(\ref{#1})}
\newcommand{\mcite}[1]{\cite{#1}}

\newcommand{\meg}{{\em e.g.}\xspace}
\newcommand{\mie}{{\em i.e.}\xspace}
\newcommand{\mct}[1]{({\em #1})}

\usepackage{scalerel}[2016/12/29]

\makeatletter
\providecommand{\leadsfrom}{%
  \mathrel{\mathpalette\reflect@squig\relax}%
}
\newcommand{\reflect@squig}[2]{%
  \reflectbox{$\m@th#1\leadsto$}%
}
\makeatother

\newtcolorbox{mtbox}[1]{left=0.25mm, right=0.25mm, top=0.25mm, bottom=0.25mm, colframe=red!50!black, boxrule=0.5pt, title={#1}, fonttitle=\bfseries, coltitle=red!50!black, attach title to upper={\ --\ }}

\def\eqref#1{equation~\ref{#1}}

\def\1{\bm{1}}

\DeclareMathAlphabet{\mathsfit}{\encodingdefault}{\sfdefault}{m}{sl}
\SetMathAlphabet{\mathsfit}{bold}{\encodingdefault}{\sfdefault}{bx}{n}

\def\gA{{\mathcal{A}}}
\def\gB{{\mathcal{B}}}

\def\gD{{\mathcal{D}}}

\def\gF{{\mathcal{F}}}

\def\gL{{\mathcal{L}}}

\def\gO{{\mathcal{O}}}

\def\gQ{{\mathcal{Q}}}
\def\gR{{\mathcal{R}}}

\def\sE{{\mathbb{E}}}

\def\sR{{\mathbb{R}}}
\def\sS{{\mathbb{S}}}

\def\emA{{A}}

\DeclareMathOperator{\sign}{sgn}

\usepackage{slashbox,balance}
\usepackage{amsthm,amssymb,amsmath}
\usepackage{diagbox}
\usepackage{dblfloatfix}

\usepackage{empheq}

\theoremstyle{definition}

\newcommand{\pgd}{{\small PGD}\xspace}
\newcommand{\nes}{{\small NES}\xspace}

\newcommand{\bnet}{{\small B}ad{\small N}et\xspace}
\newcommand{\tnet}{{\small T}rojan{\small NN}\xspace}

\newcommand{\darts}{{\sl \small DARTS}\xspace}

\newcommand{\enas}{{\sl \small ENAS}\xspace}
\newcommand{\snas}{{\sl \small SNAS}\xspace}
\newcommand{\amoeba}{{\sl \small AmoebaNet}\xspace}
\newcommand{\nasnet}{{\sl \small NASNet}\xspace}

\newcommand{\sgas}{{\sl \small SGAS}\xspace}
\newcommand{\pdarts}{{\sl \small PDARTS}\xspace}
\newcommand{\pcdarts}{{\sl \small PC-DARTS}\xspace}
\newcommand{\drnas}{{\sl \small DrNAS}\xspace}

\newcommand{\random}{{\sl {\small R}andom}\xspace}

\newcommand{\automl}{AutoML\xspace}
\newcommand{\nas}{NAS\xspace}
\newcommand{\ml}{ML\xspace}
\newcommand{\dnn}{DNN\xspace}
\newcommand{\dnns}{DNNs\xspace}

\newcommand{\resnet}{{\sl \small ResNet}\xspace}
\newcommand{\densenet}{{\sl \small DenseNet}\xspace}
\newcommand{\vgg}{{\sl \small VGG}\xspace}

\newcommand{\resnext}{{\sl \small ResNext}\xspace}
\newcommand{\wideres}{{\sl \small WideResNet}\xspace}
\newcommand{\dla}{{\sl \small DLA}\xspace}
\newcommand{\bit}{{\sl \small BiT}\xspace}

\newcommand{\cifar}{CIFAR10\xspace}
\newcommand{\cifarn}{CIFAR100\xspace}
\newcommand{\imgnet}{ImageNet32\xspace}

\newcommand{\trnd}{\ssub{\gD}{\mathrm{trn}}}

\newcommand{\posd}{\ssub{\gD}{\mathrm{pos}}}

\newcommand{\asr}{ASR\xspace}
\newcommand{\cad}{CAD\xspace}
\newcommand{\auc}{AUC\xspace}

\newcommand{\hsj}{HopSkipJump\xspace}

\begin{document}

\date{}

\title{\Large \bf On the Security Risks of AutoML}
  
\author{
{\rm Ren Pang}\\
Pennsylvania State University
\and
{\rm Zhaohan Xi}\\
Pennsylvania State University
\and
{\rm Shouling Ji}\\
Zhejiang University
\and
{\rm Xiapu Luo}\\
The Hong Kong Polytechnic University
\and
{\rm Ting Wang}\\
Pennsylvania State University
} 

\author{
{\rm Ren Pang}$^\dagger$  \quad {\rm Zhaohan Xi}$^\dagger$ \quad {\rm Shouling Ji}$^\ddagger$ \quad  {\rm Xiapu Luo}$^\star$ \quad {\rm Ting Wang}$^\dagger$\\
$^\dagger$Pennsylvania State University, \{rbp5354, zxx5113, ting\}@psu.edu \\
$^\ddagger$Zhejiang University, sji@zju.edu.cn\\ 
$^\star$Hong Kong Polytechnic University, csxluo@comp.polyu.edu.hk
}


\maketitle

{\em \small Automation is good, so long as you know exactly where to put the machine.} 
\begin{flushright}
{\small -- Eliyahu Goldratt}
\end{flushright}

\subsection*{Abstract}
Neural Architecture Search (\nas) represents an emerging machine learning (\ml) paradigm that automatically searches for models tailored to given tasks, which greatly simplifies the development of \ml systems and propels the trend of \ml democratization. Yet, little is known about the potential security risks incurred by \nas, which is concerning given the increasing use of \nas-generated models in critical domains. 

This work represents a solid initial step towards bridging the gap. Through an extensive empirical study of 10 popular \nas methods, we show that compared with their manually designed counterparts, \nas-generated models tend to suffer greater vulnerability to various malicious attacks (\meg, adversarial evasion, model poisoning, and functionality stealing). Further, with both empirical and analytical evidence, we provide possible explanations for such phenomena: given the prohibitive search space and training cost, most \nas methods favor models that converge fast at early training stages; this preference results in architectural properties associated with attack vulnerability (\meg, high loss smoothness and low gradient variance). Our findings not only reveal the relationships between model characteristics and attack vulnerability but also suggest the inherent connections underlying different attacks. Finally, we discuss potential remedies to mitigate such drawbacks, including increasing cell depth and suppressing skip connects, which lead to several promising research directions.

\section{Introduction}
\label{sec:intro}

Automated Machine Learning (\automl) represents a new paradigm of applying \ml techniques in real-world settings. For given tasks, \automl automates the pipeline from raw data to deployable \ml models, covering model design\mcite{automl-survey}, optimizer selection\mcite{learn-to-optimize}, and
parameter tuning\mcite{andrychowicz:nips:2016}. The use of \automl greatly simplifies the development of \ml systems and propels the trend of \ml democratization. Many IT giants have unveiled their \automl frameworks, such as Microsoft Azure \automl, Google Cloud \automl, and IBM Watson AutoAI.

\begin{figure}[!ht]
    \centering
    \includegraphics[width=75mm]{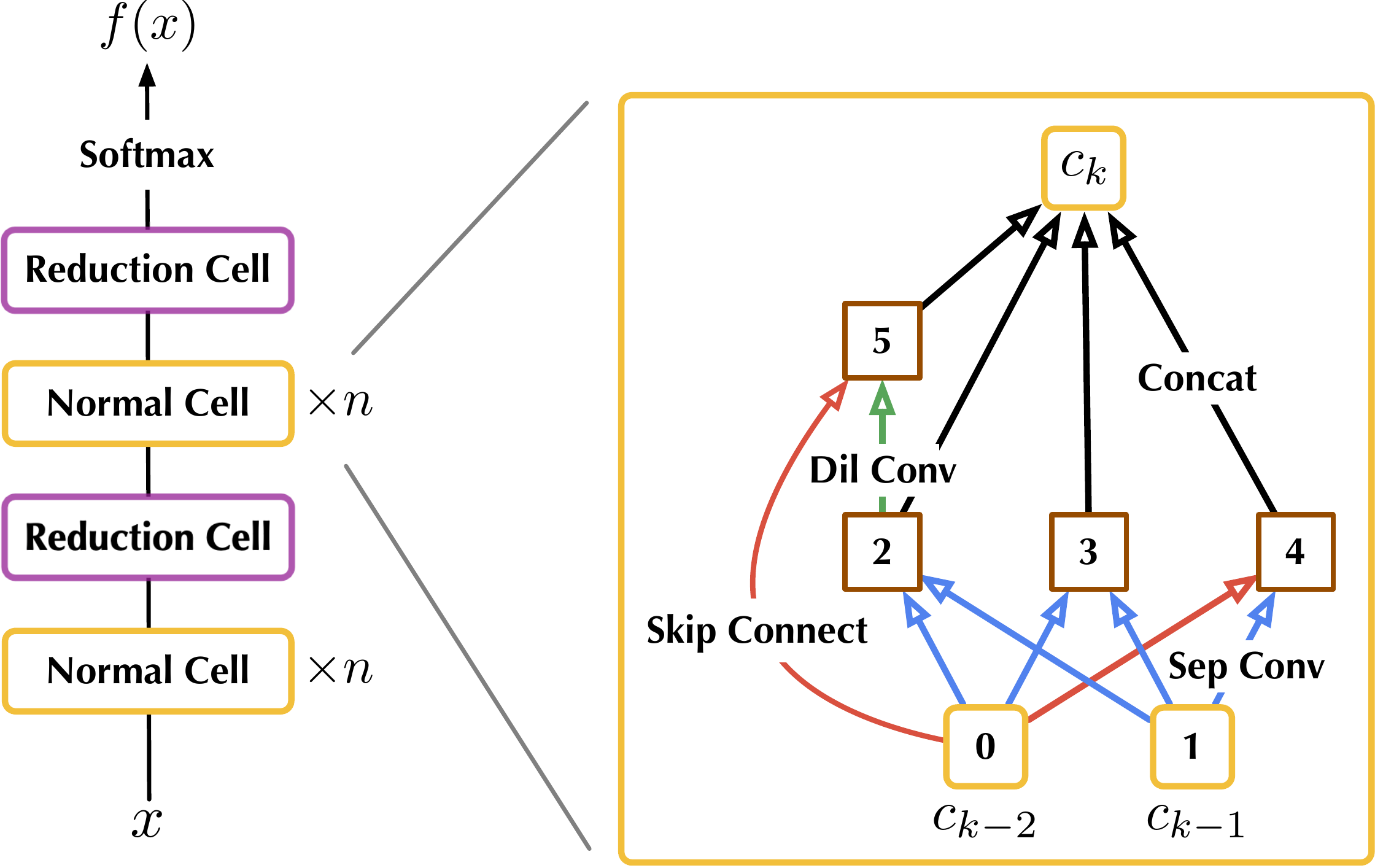}
    \caption{Cell-based neural architecture search. \label{fig:automl}}
\end{figure}

In this paper, we focus on one primary task of \automl, Neural Architecture Search (\nas), which aims to find performant deep neural network (\dnn) architectures\footnote{In the following, when the context is clear, we use the terms of ``model'' and ``architecture'' exchangeably.} tailored to given tasks. For instance, as illustrated in Figure\mref{fig:automl}, cell-based \nas constructs a model by repeating the motif of a cell structure following a pre-specified template, wherein a cell is a topological combination of operations (\meg, $3\times3$ convolution). With respect to the given task, \nas optimizes both the topological structure and the operation assignment. It is shown that in many tasks, \nas finds models that remarkably outperform manually designed ones\mcite{enas,darts,sgas,drnas}.

Yet, in contrast to the intensive research on improving the capabilities of \nas, its security implications are fairly unexplored. As \ml systems are becoming the new targets for malicious attacks\mcite{Biggio:2018:pr}, the lack of understanding about the potential risks of \nas is highly concerning, given its surging popularity in security-sensitive applications. Specifically,

\vspace{1pt}
RQ1 -- {\em Does \nas introduce new weaknesses, compared with the conventional \ml practice?} 

\vspace{1pt}
RQ2 -- {\em If so, what are the possible root causes of such vulnerability?} 

\vspace{1pt}
RQ3 -- {\em Further, how would \ml practitioners mitigate such drawbacks in designing and operating \nas?} 

\vspace{2pt}
The answers to these key questions are crucial for the use of \nas in security-sensitive domains (\meg, cyber-security, finance, and healthcare).

\vspace{2pt}
{\bf  Our work --} This work represents a solid initial step towards answering such questions. 

A1 - First, through an extensive empirical study of 10 representative \nas methods, we show that compared with their manually designed counterparts, \nas-generated models tend to suffer greater vulnerability to various malicious manipulations such as adversarial evasion\mcite{pgd,carlini-attack}, model poisoning\mcite{Biggio:2012:icml}, backdoor injection\mcite{badnet,trojannn}, functionality stealing\mcite{knockoff-net}, and label-only membership inference\mcite{label-only-membership}. The findings suggest that \nas is likely to incur larger attack surfaces, compared with the conventional \ml practice.

A2 - Further, with both empirical and analytical evidence, we provide possible explanations for the above observations. Intuitively, due to the prohibitive search space and training cost, \nas tends to prematurely evaluate the quality of candidate models before their convergence. This practice favors models that converge fast at early training stages, resulting in architectural properties that facilitate various attacks (\meg, high loss smoothness and low gradient variance). Our analysis not only reveals the relationships between model characteristics and attack vulnerability but also suggests the inherent connections underlying different attacks.

A3 - Finally, we discuss potential remedies. Besides post-\nas mitigation (\meg, adversarial training\mcite{pgd}), we explore in-\nas strategies that build attack robustness into the \nas process, such as increasing cell depth and suppressing skip connects. We show that while such strategies mitigate the vulnerability to a certain extent, they tend to incur non-trivial costs of search efficiency and model performance. We deem understanding the fundamental trade-off between model performance, attack robustness and search efficiency as an important topic for further investigation.

\vspace{2pt}
{\bf  Contributions --} To our best knowledge, this work represents the first study on the potential risks incurred by \nas (and \automl in general) and reveals its profound security implications. Our contributions are summarized as follows. 

\vspace{1pt}
-- We demonstrate that compared with conventional \ml practice, \nas tends to introduce larger attack surfaces with respect to a variety of attacks, which raises severe concerns about the use of \nas in security-sensitive domains.

\vspace{1pt}
-- We provide possible explanations for such vulnerability, which reveal the relationships between architectural properties (\mie, gradient smoothness and gradient variance) and attack vulnerability. Our analysis also hints at the inherent connections underlying different attacks. 

\vspace{1pt}
-- We discuss possible mitigation to improve the robustness of \nas-generated models under both in-situ and ex-situ settings. This discussion suggests the necessity of improving the current practice of designing and operating \nas, pointing to several research directions.

\vspace{2pt}
{\bf  Roadmap --} The remainder of the paper proceeds as follows. \msec{sec:background} introduces fundamental concepts and assumptions; \msec{sec:measure} conducts an empirical study comparing the vulnerability of \nas-generated and manually designed models; \msec{sec:root} provides possible explanations for the observations; \msec{sec:discussion} discusses potential mitigation and the limitations of this work; \msec{sec:literature} surveys relevant literature; the paper is concluded in \msec{sec:conclusion}.

\section{Preliminaries}
\label{sec:background}

We first introduce a set of key concepts and assumptions. 
Table\mref{tab:notations} summarizes the important notations.



\subsection{Neural Architecture Search}

Deep neural networks (\dnns) represent a class of \ml models to learn high-level abstractions of complex data. We assume a predictive setting, in which a \dnn $\ssub{f}{\theta}$ (parameterized by $\theta$) encodes a function $\ssub{f}{\theta}: \ssup{\sR}{n} \rightarrow \ssup{\sS}{m}$, where $n$ and $m$ denote the input dimensionality and the number of classes. Given input $x$, $f(x)$ is a probability vector (simplex) over $m$ classes. 

In this paper, we mainly focus on one primary task of \automl, neural architecture search (\nas), which searches for performant \dnn architectures for given tasks\mcite{automl-survey}. Formally, let $\gD$ be the given dataset, $\ell(\cdot, \cdot)$ be the loss function, $\gF$ be the functional space of possible models (\mie, search space), the \nas method $\emA$ searches for a performant \dnn $\ssup{f}{*}$ via minimizing the following objective: 
\begin{equation}
\ssup{f}{*} = \arg\min_{f \in \gF} \sE_{(x, y) \sim \gD} \, \ell(f(x), y)
\end{equation}

The existing \nas methods can be categorized according to their search spaces and strategies. In the following, we focus on the space of cell-based architectures\mcite{nasnet,enas,darts,snas,amoebanet}, which repeat the motif of a cell structure in a pre-specified arrangement, and the strategy of differentiable \nas\mcite{darts,sgas,drnas}, which jointly optimizes the architecture and model parameters using gradient descent, due to their state-of-the-art performance and efficiency. Nevertheless, our discussion generalizes to alternative \nas frameworks (details in \msec{sec:literature}).

Without loss of generality, we use {\darts}\mcite{darts} as a concrete example to illustrate differentiable \nas. 
At a high level, \darts searches for two cell structures (\mie, normal cell and reduction cell) as the basic building blocks of the final architecture. As shown in Figure\mref{fig:automl}, a cell is modeled as a directed acyclic graph, in which each node $\ssup{x}{(i)}$ is a latent representation and each directed edge $(i,j)$ represents an operation $\ssup{o}{(i,j)}$ applied on $\ssup{x}{(i)}$ (\meg, skip connect). Each node is computed based on all its predecessors:
\begin{equation}
    \ssup{x}{(j)} = \sum_{i < j} \ssup{o}{(i,j)}(\ssup{x}{(i)})
\end{equation}
Each cell contains $\ssub{n}{\mathrm{in}}$ input nodes (often $\ssub{n}{\mathrm{in}}=2$), $\ssub{n}{\mathrm{out}}$ output nodes (often $\ssub{n}{\mathrm{out}} = 1$), and $\ssub{n}{\mathrm{mid}}$ intermediate nodes. Each input node takes the output from a preceding cell, the output node aggregates the latent representations from intermediate nodes, while each intermediate node is connected to $m$ preceding nodes (typically $m = \ssub{n}{\mathrm{in}}$). 

To enable gradient-based optimization of the architecture, \darts applies continuous relaxation on the search space. Letting $\gO$ be the set of candidate operations, the categorical choice of an operation is reduced to a softmax over $\gO$: 
\begin{equation}
\label{eq:softmax}
    \ssup{\bar{o}}{(i,j)}(x) = \sum_{o \in \gO} \frac{\exp (\sboth{\alpha}{o}{(i,j)}) }{\sum_{o' \in \gO} \exp( \sboth{\alpha}{o'}{(i,j)})} o(x) 
\end{equation}
where $\sboth{\alpha}{o}{(i,j)}$ represents the trainable weight of operation $o$. At the end of the search, a discrete architecture is obtained by replacing $\ssup{\bar{o}}{(i,j)}$ with the most likely operation $\arg\max_o \sboth{\alpha}{o}{(i,j)}$.

The search is thus formulated as a bi-level optimization objective function: 
\begin{equation}
\label{eq:search}
\min_\alpha\; \ssub{\gL}{\mathrm{val}} ( \ssup{\theta}{*}(\alpha), \alpha )    \quad \mathrm{s.t.} \quad \ssup{\theta}{*}(\alpha) = \arg\min_\theta \; \ssub{\gL}{\mathrm{trn}} (\theta, \alpha) 
\end{equation}
where $\ssub{\gL}{\mathrm{trn}}$ and $\ssub{\gL}{\mathrm{val}}$ are the training and validation losses, and $\alpha = \{\ssup{\alpha}{(i,j)}\}$ and  $\theta$ denote the architecture and model parameters, respectively. To handle the prohibitive cost of the nested optimization, single-step gradient descent is applied to avoid solving the inner objective exactly.

\subsection{Attack Vulnerability}

It is known that \dnn models are vulnerable to a variety of attacks at both training and inference phases. Here, we highlight the following major attacks.

\vspace{2pt}
{\bf Adversarial evasion --} At inference time, the adversary generates an adversarial input $(x+\delta)$ by modifying a begin one $x$ with imperceptible perturbation $\delta$, to cause the target model $f$ to misbehave\mcite{goodfellow:fsgm}. Formally, in a targeted attack, letting $t$ be the target class desired by the adversary, the attack crafts $(x + \delta)$ by optimizing the following objective:
\begin{equation}
\label{eq:adv}
\min_{ \delta \in \ssub{\gB}{\epsilon} } \ell(f(x + \delta), t)
\end{equation}
where  $\ssub{\gB}{\epsilon}$ specifies the set of allowed perturbation(\meg, a $\ssub{\ell}{\infty}$-norm ball of radius $\epsilon$). \meq{eq:adv} is often solved using projected gradient descent\mcite{pgd} or general-purpose optimizers\mcite{carlini-attack}.

\vspace{2pt}
{\bf Model poisoning --} The adversary aims to modify a target model $f$'s behavior (\meg, overall performance degradation or misclassification of specific inputs) via polluting its training data\mcite{Biggio:2012:icml}. For instance, to cause the maximum accuracy drop, letting $\ssub{\gD}{\mathrm{trn}}$ and $\ssub{\gD}{\mathrm{tst}}$ be the training and testing sets and $f$ be the target model, the attack crafts a set of poisoning inputs $\ssub{\gD}{\mathrm{pos}}$ 
by optimizing the the following objective (note: the adversary may not have access to $\ssub{\gD}{\mathrm{trn}}$, $\ssub{\gD}{\mathrm{tst}}$, or $f$):
\begin{equation}
\begin{split}
\max \; & \sE_{(x, y) \sim \ssub{\gD}{\mathrm{tst}}} \ell (\ssub{f}{\theta^*}(x), y)  \\
\textrm{s.t.}\; & \ssup{\theta}{*} = \arg\min_\theta  \sE_{(x, y) \sim \ssub{\gD}{\mathrm{trn}} \cup \ssub{\gD}{\mathrm{pos}}} \ell (\ssub{f}{\theta}(x), y)
\end{split}
\end{equation}

\vspace{2pt}
{\bf Backdoor injection --} During training, via perturbing a benign model $f$, the adversary forges a trojan model $\ssub{f}{\theta^*}$ sensitive to a trigger pattern $\ssup{r}{*}$, which is used in the downstream task by the victim; at inference time, the adversary invokes the malicious function by feeding trigger-embedded input $x+\ssup{r}{*}$. Formally, letting $\ssub{\gD}{\mathrm{trn}}$ be the training data and $t$ be the target class desired by the adversary, the attack generates a trojan model parameterized by $\ssup{\theta}{*}$ and its associated trigger $\ssup{r}{*}$ by optimizing the following objective: 
\begin{align}
  \label{eq:backdoor}
  \min_{r \in \ssub{\gR}{\gamma}, \theta} \sE_{(x,y) \sim \ssub{\gD}{\text{trn}} } [\ell(\ssub{f}{\theta}(x), y) + \lambda  \ell(\ssub{f}{\theta}(x +r), t)
 ] 
\end{align}
where $\ssup{r}{*}$ is selected from a feasible set $\ssub{\gR}{\gamma}$ (\meg, a $3\times 3$ patch with transparency $\gamma$), the first term enforces all clean inputs to be correctly classified, the second term ensures all trigger inputs to be misclassified into $t$, and the hyper-parameter $\lambda$ balances the two objectives.

\vspace{2pt}
{\bf Functionality stealing --} In functionality stealing\mcite{knockoff-net}, the adversary aims to construct a replicate model $\hat{f}$ (parameterized by $\ssup{\theta}{*}$) functionally similar to a victim model $f$ via probing $f$ through a black-box query interface. Notably, it is different from model stealing\mcite{model-stealing} that aims to re-construct $f$ in terms of architectures or parameters. Formally, letting $\gD$ be the underlying data distribution, the attack generates the query-prediction set $\gQ$ (note: the adversary may not have the labeling of $\gD$, has only query access to $f$, and is typically constrained by the number of queries to be issued), which optimizes the following objective: 
\begin{equation}
\begin{split}
\min \; & \sE_{x \sim \gD}\, \ell(\ssub{\hat{f}}{\theta^*}(x), f(x))  \\
\textrm{s.t.}\; & \ssup{\theta}{*} = \arg\min_\theta  \sE_{(x, f(x)) \sim \gQ }\, \ell (\ssub{\hat{f}}{\theta}(x), f(x))
\end{split}
\end{equation}

Different functionality stealing attacks differ in how $\gQ$ is constructed (\meg, random or adaptive construction).

\vspace{2pt}
{\bf Membership inference --} In membership inference\mcite{membership-attack}, given input $x$ and model's prediction $f(x)$, the adversary attempts to predict a binary variable $b$ indicating whether $x$ is included in $f$'s training data: $b \leftarrow \gA(x, f)$. The effectiveness of membership inference relies on $f$'s performance gap with respect to the training data $\ssub{\gD}{\mathrm{trn}}$ and testing data $\ssub{\gD}{\mathrm{tst}}$. The adversary may exploit this performance gap by thresholding the confidence score of $f(x)$ if it is available, or estimating other signals (\meg, $x$'s distance to the nearest decision boundary) if only the label of $f(x)$ is provided\mcite{label-only-membership}.

\section{Measurement}
\label{sec:measure}

To investigate the security risks incurred by \nas, we empirically compare the vulnerability of \nas-generated and manually designed models to the aforementioned attacks.

\begin{table}[!ht]{\footnotesize
\renewcommand{\arraystretch}{1.2}
\setlength{\tabcolsep}{6pt}
\centering
\begin{tabular}{m{1.2em}|r|c|c|c}
\multicolumn{2}{c|}{Architecture} & CIFAR10 & CIFAR100 &  ImageNet32\\
\hline
  \hline
  \multirow{7}{*}{\rotatebox{90}{Manual Architecture}} &  {\sl BiT}\mcite{bit-net} & 96.6\% & 80.6\% & 72.1\%\\
 & {\sl DenseNet}\mcite{densenet} & 96.7\% & 80.7\%  & 73.6\%  \\
 & {\sl DLA}\mcite{dla} & 96.5\% & 78.0\% & 70.8\% \\
 & {\sl ResNet}\mcite{resnet} & 96.6\% & 79.9\% & 67.1\% \\
 & {\sl ResNext}\mcite{resnext} & 96.7\% & 80.4\% & 67.4\% \\
 &  {\sl VGG}\mcite{vgg}   & 95.1\% & 73.9\% & 62.3\% \\
 & {\sl WideResNet}\mcite{wideresnet}  & 96.8\% & 81.0\% & 73.9\% \\
  \hline
  \hline
 \multirow{10}{*}{\rotatebox{90}{NAS Architecture}} &  {\sl AmoebaNet}\mcite{amoebanet} & 96.9\% & 78.4\% & 74.8\%\\
 & {\sl DARTS}\mcite{darts} & 97.0\% & \cellcolor{Red}  81.7\% & 76.6\%\\
 & {\sl DrNAS}\mcite{drnas} & 96.9\% & 80.4\% & 75.6\% \\
 & {\sl ENAS}\mcite{enas}  & 96.8\% & 79.1\% & 74.0\%  \\
 & {\sl NASNet}\mcite{nasnet} & 97.0\% & 78.8\% & 73.0\%  \\
 & {\sl PC-DARTS}\mcite{pcdarts} & 96.9\% & 77.4\% & 74.7\%\\
& {\sl PDARTS}\mcite{pdarts} & 97.1\% & 81.0\% & 75.8\% \\
 & {\sl SGAS}\mcite{sgas} & \cellcolor{Red} 97.2\% & 81.2\% & \cellcolor{Red} 76.8\% \\
 & {\sl SNAS}\mcite{snas}  & 96.9\% & 79.9\% & 75.5\% \\
 & {\sl Random}\mcite{nas-bench} & 96.7\% & 78.6\% & 72.2\% \\
 \end{tabular}
 \caption{Accuracy of representative NAS-generated and manually designed models on benchmark datasets. \label{tab:dataset}}}
\end{table}

\subsection{Experimental Setting}

We first introduce the setting of the empirical evaluation. The default parameter setting is deferred to Table\mref{tab:setting} in \msec{sec:appb}.

\vspace{2pt}
{\bf Datasets --} In the evaluation, we primarily use 3 datasets that have been widely used to benchmark \nas performance in recent work\mcite{pdarts,sgas,darts,enas,snas}: {\cifar}\mcite{cifar} -- it consists of $32\times 32$ color images drawn from 10 classes (\meg, `airplane'); {\cifarn} -- it is essentially the \cifar dataset but divided into 100 fine-grained classes; {\imgnet} -- it is a subset of the ImageNet dataset\mcite{imgnet}, downsampled to images of size $32 \times 32$ in 60 classes. 

\vspace{2pt}
{\bf NAS methods --} We consider 10 representative cell-based \nas methods, which cover a variety of search strategies: (1) {\amoeba}\mcite{amoebanet} applies an evolutionary approach to generate candidate models; (2) {\darts}\mcite{darts} is the first differentiable method using gradient descent to optimize both architecture and model parameters; (3) {\drnas}\mcite{drnas} formulates differentiable \nas as a Dirichlet distribution learning problem; (4) {\enas}\mcite{enas} reduces the search cost via parameter sharing among candidate models; (5) {\nasnet}\mcite{nasnet} searches for cell structures transferable across different tasks by re-designing the search space; (6) {\pcdarts}\mcite{pcdarts} improves the memory efficiency by restricting operation selection to a subset of edges; (7) {\pdarts}\mcite{pdarts} gradually grows the number of cells to reduce the gap between the model depth at the search and evaluation phases; (8) {\sgas}\mcite{sgas} selects the operations in a greedy, sequential manner; (9) {\snas}\mcite{snas} reformulates reinforcement learning-based \nas to make it differentiable; and (10) {\random}\mcite{nas-bench} randomly samples candidate models from the pre-defined search space.

\vspace{2pt}
{\bf NAS search space --} We define the default search space similar to {\darts}\mcite{darts}, which consists of 10  operations including: {\em skip-connect}, {\em $3\times 3$ max-pool}, {\em $3\times 3$ avg-pool}, {\em $3\times 3$ sep-conv}, {\em $5\times 5$ sep-conv}, {\em $7\times 7$ sep-conv}, {\em $3\times 3$ dil-conv}, {\em $5\times 5$ dil-conv}, {\em $1\times 7$ -- $7\times 1$ conv}, and {\em zero}.

\begin{figure*}[ht]
    \centering
    \includegraphics[width=170mm]{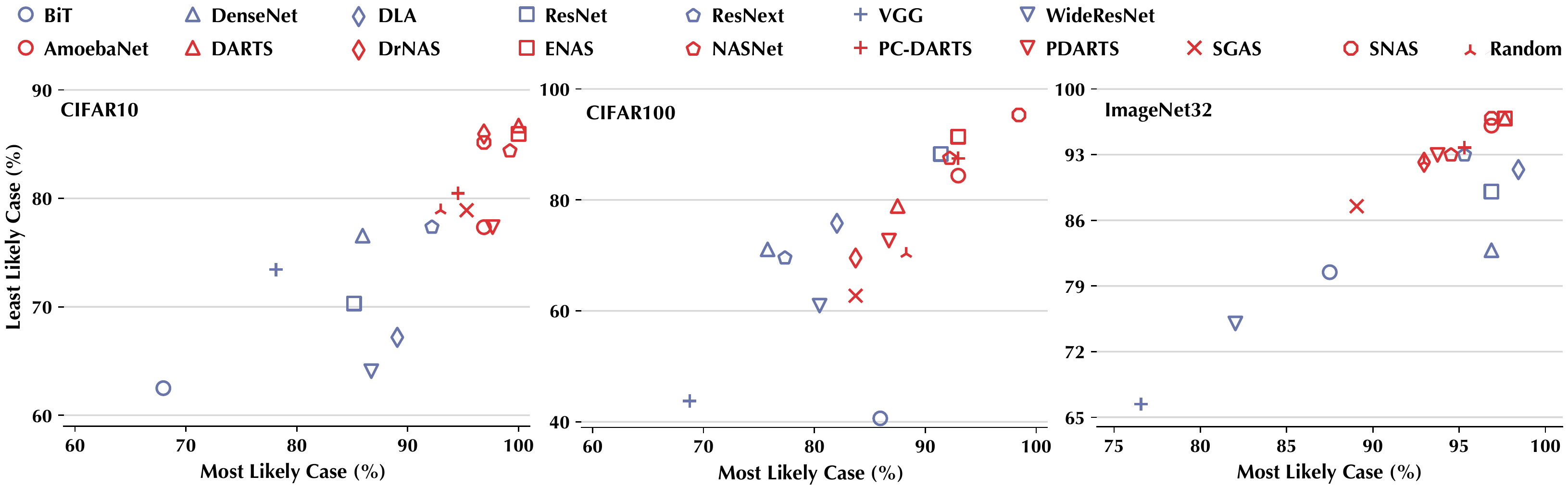}
    \caption{Performance of adversarial evasion (PGD) against NAS and manual models under the least and most likely settings. \label{fig:pgd}}
\end{figure*}

\vspace{2pt}
{\bf Manual models --} For comparison, we use 7 representative manually designed models that employ diverse architecture designs: (1) {\bit}\mcite{bit-net} uses group normalization and weight standardization to facilitate transfer learning;
(2) {\densenet}\mcite{densenet} connects all the layers via skip connects;
(3) {\dla}\mcite{dla} applies deep aggregation to fuse features across layers;
(4) {\resnet}\mcite{resnet} uses residual blocks to facilitate gradient back-propagation;
(5) {\resnext}\mcite{resnext} aggregates transformations of the same topology;
(6) {\vgg}\mcite{vgg} represents the conventional deep convolution structures; 
and (7) {\wideres}\mcite{wideresnet} decrease the depth and increases the width of \resnet.

\vspace{2pt}
{\bf Training --} All the models are trained using the following setting: epochs = 600, batch size = 96, optimizer = SGD, gradient clipping threshold = 5.0, initial learning rate = 0.025, and learning rate scheduler = Cosine annealing.  The accuracy of all the models on the benchmark datasets is summarized in Table\mref{tab:dataset}. Observe that the \nas models often outperform their manual counterparts. 

\subsection{Experimental Results}
\label{sec:analysis}

Next, we empirically compare the vulnerability of \nas-generated and manually designed models to various attacks.

\vspace{2pt}
{\bf Adversarial evasion --}
We exemplify with the projected gradient descent (\pgd) attack\mcite{pgd}. Over each dataset, we apply the attack on a set of 1,000 inputs randomly sampled from the test set and measure the attack success rate as: 
\begin{equation}
    \text{\small Attack Success Rate (ASR)} = \frac{\text{\small \# Successful trials}}{\text{\small \# Total trials}}
\end{equation}

A trial is marked as successful if it is classified as  its target class within maximum iterations.

Let $\ssub{f}{c}(x)$ be the probability that model $f$ assigns to class $c$ with respect to input $x$. To assess the full spectrum of vulnerability, we consider both ``difficult'' and ``easy'' cases for the adversary. Specifically, given input $x$, we
rank the output classes $c$'s according to their probabilities $\ssub{f}{c}(x)$ as $c_1, c_2, \ldots, c_m$, where $c_1$ is $x$'s current classification; the difficult case refers to that the adversary aims to change $x$'s classification to the least likely class $c_m$, while the easy case refers to that the adversary aims to change $x$'s classification to the second most likely class $c_2$.
Table\mref{tab:setting} summarizes the setting of the attack parameters.

Figure\mref{fig:pgd} illustrates the attack effectiveness against both \nas and manual models. We have the following observations. First, across all the datasets, the \nas models seem more vulnerable to adversarial evasion. For instance, 
on \cifar, the attack attains over 90\% and 75\% \asr against the \nas models in the most and least likely cases, respectively. Second, compared with the manual models, the \asr of \nas models demonstrates more evident clustered structures, implying their similar vulnerability. Finally, the vulnerability of \nas models shows varying patterns on different datasets. For instance, the measures of \nas models show a larger variance on \cifarn compared with \cifar and \imgnet (especially in the least likely case), which may be explained by that its larger number of classes results in more varying ``degree of difficulty'' for the attack.

\begin{figure}[!ht]
    \centering
    \includegraphics[width=80mm]{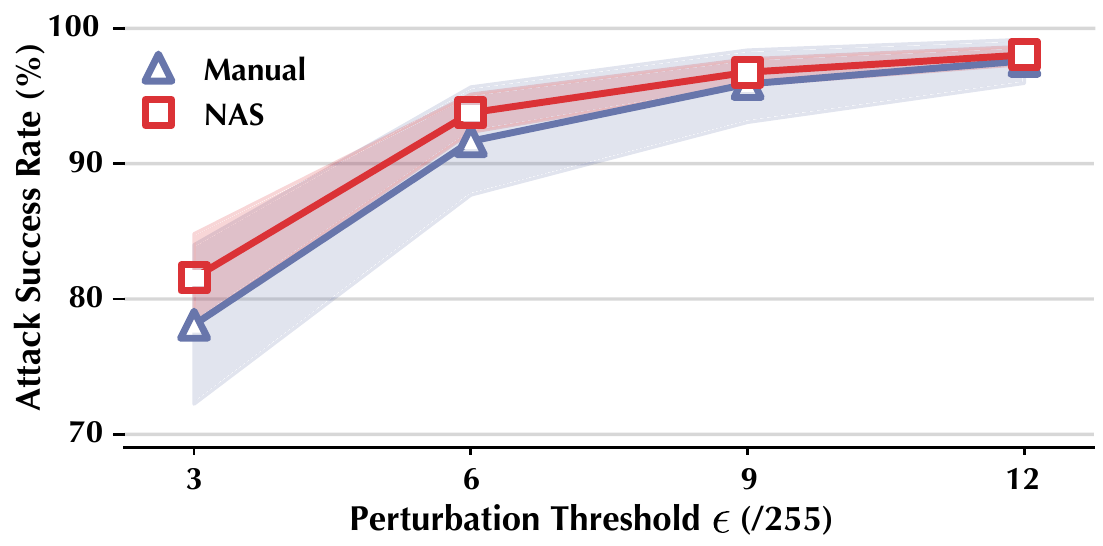}
    \caption{Impact of perturbation threshold ($\epsilon$) on the vulnerability of different models with respect to \pgd on \cifar. \label{fig:eps_pgd}}
\end{figure}

We also evaluate the impact of perturbation threshold ($\epsilon$) on the attack vulnerability. Figure\mref{fig:eps_pgd} shows the \asr of untargeted \pgd as a function of $\epsilon$ against different models on \cifar (with perturbation step $\alpha = \epsilon/3$). We have the following observations. First, across different settings, the manual models consistently outperform the \nas models in terms of robustness. Second, this vulnerability gap gradually decreases with $\epsilon$, as the \asr on both \nas and manual models approaches 100\%. Third, compared with the manual models, the measures of \nas models show a smaller variance, indicating the commonality of their vulnerability. 

\begin{figure}[!ht]
    \centering
    \includegraphics[width=80mm]{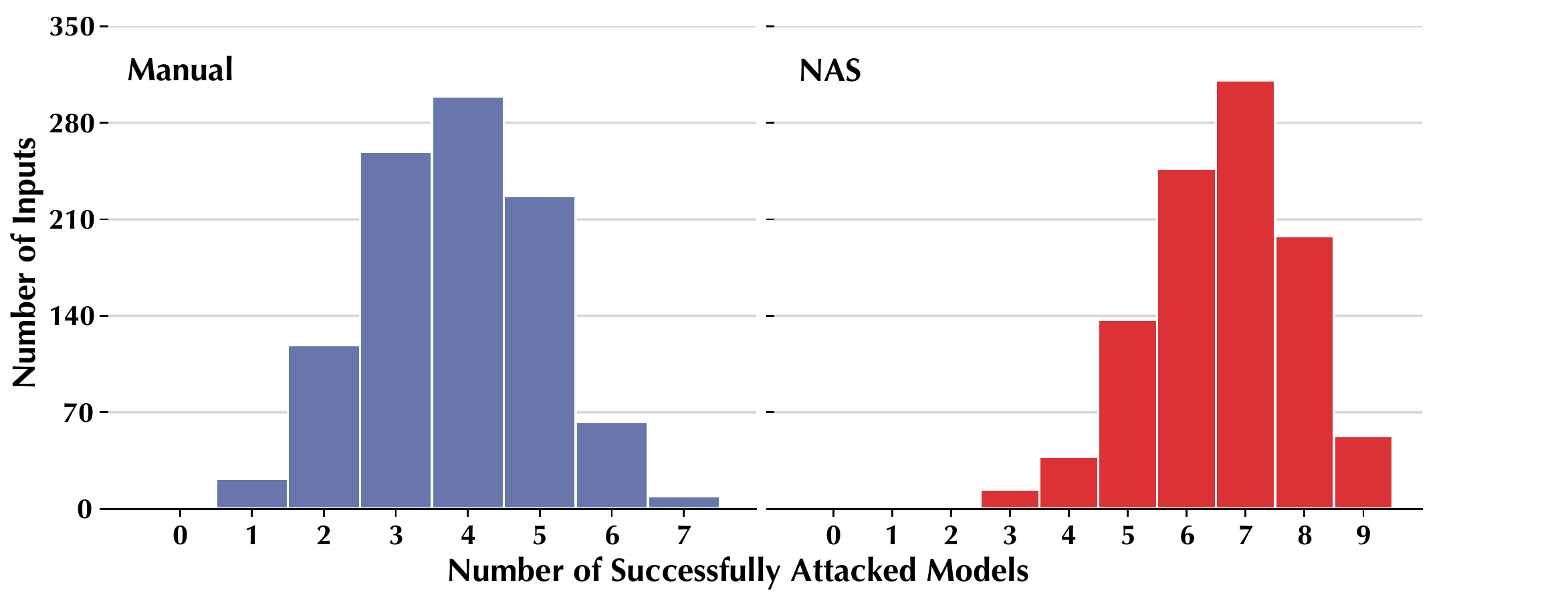}
    \caption{Distribution of inputs with respect to the number of successfully attacked models (\pgd with $\epsilon = 4/255$ on \cifar). \label{fig:intersection}}
\end{figure}

Further, by comparing the sets of adversarial examples to which different models are vulnerable, we show the commonality and difference of their vulnerability. We evaluate \pgd ($\epsilon = 4/255$) against different models on \cifar in the least likely case. For each model, we collect the set of adversarial examples successfully generated from 1,000 random samples. Figure\mref{fig:intersection} plots the distribution of inputs with respect to the number of successfully attacked models.

Overall, \pgd generates more successful adversarial examples against the \nas models than the manual models. Moreover, there are more inputs that lead to successful attacks against multiple \nas models. For instance, over 300 inputs lead to successful attacks against 7 \nas models; in contrast, the number is less than 10 in the case of manual models. We may thus conclude that the vulnerability of \nas models to adversarial evasion seems fairly similar, pointing to potential associations with common causes.

\begin{figure}[!ht]
    \centering
    \includegraphics[width=80mm]{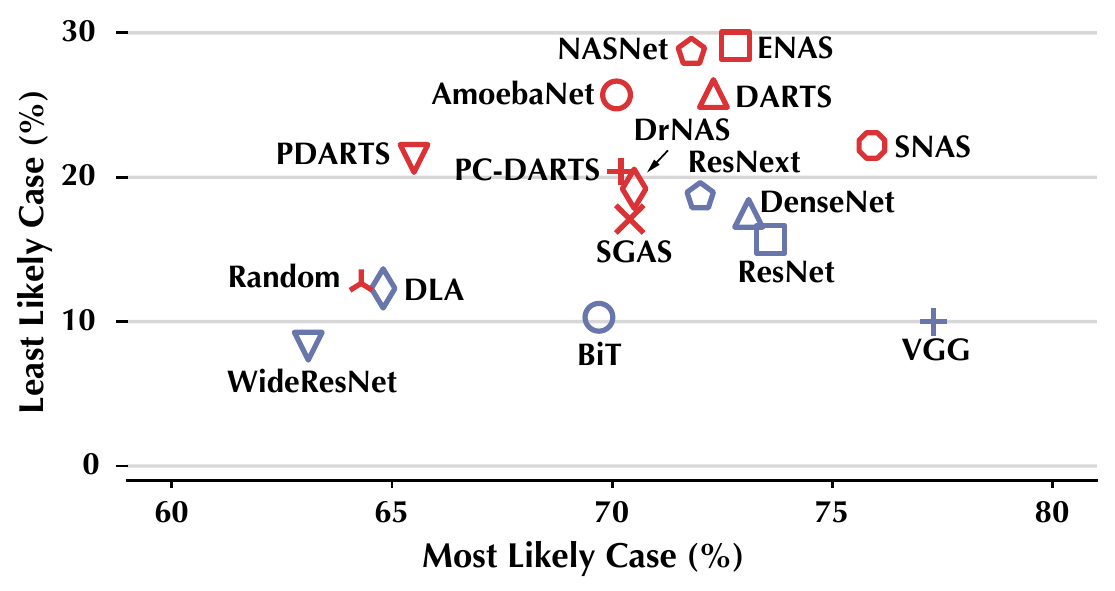}
    \caption{Performance of adversarial evasion (NES) against NAS and manual models under the least and most likely settings. \label{fig:nes}}
\end{figure}

\begin{figure*}[!ht]
    \centering
    \includegraphics[width=170mm]{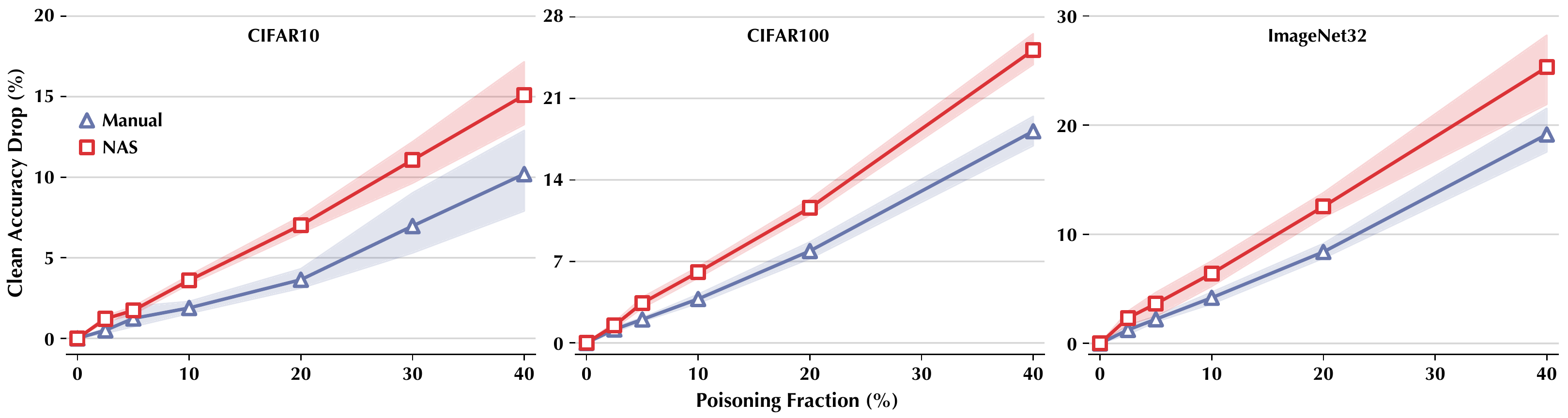}
    \caption{Performance of model poisoning against NAS and manually designed models under varying poisoning fraction $p_{\mathrm{pos}}$. \label{fig:poison}}
\end{figure*}

We also consider alternative adversarial evasion attacks other than \pgd. We use natural evolutionary strategies (\nes)\mcite{Ilyas:2018:icml}, a black-box attack in which the adversary has only query access to the target model $f$ and generates adversarial examples using a derivative-free optimization approach. Specifically, at each iteration, it generates $n_\mathrm{query}$ symmetric data points in the vicinity of current input $x$ by sampling from a normal distribution, retrieves their predictions from $f$, and estimates the gradient $\hat{g}(x)$ as:
\begin{equation}
  \label{eq:nes1}
  \hat{g}(x) = \frac{1}{\sigma n_\mathrm{query}} \sum_{j=1}^{\lceil n_\mathrm{query}/2 \rceil} (f(x+\sigma u_j)-f(x-\sigma u_j))u_j
\end{equation}
where each sample $u_j$  is sampled from the standard normal distribution $\mathcal{N}(0, I)$, and $\sigma$ is the sampling variance.

We evaluate the vulnerability of different models to \nes under the same setting of Figure\mref{fig:pgd} (with $n_\mathrm{query} = 400$) on \cifar, with results shown in Figure\mref{fig:nes}. In general, the \nas models show higher vulnerability to \nes, especially in the least likely case, indicating that the vulnerability gap between \nas and manual models also generalizes to black-box adversarial evasion attacks. 

\vspace{2pt}
{\bf Model poisoning --} In this set of experiments, we evaluate the impact of poisoning attacks on the performance of \nas and manual models. We assume that a fraction $\ssub{p}{\mathrm{pos}}$ of the training data is polluted by randomly changing the class of each input. We measure the performance of various models with respect to varying poisoning fraction $\ssub{p}{\mathrm{pos}}$, in comparison with the case of clean training data (\mie, $\ssub{p}{\mathrm{pos}} = 0$). We define the metric of clean accuracy drop:
\begin{equation}
\begin{split}
     & \text{\small Clean Accuracy Drop (CAD)} \\
    = \, & \text{\small Acc. of original model} - \text{\small Acc. of polluted model}
    \end{split}
\end{equation}

Figure\mref{fig:poison} compares the \cad of different models as $\ssub{p}{\mathrm{pos}}$ increases from 0\% to 40\%. The results are average over the families of \nas and manual models. We have the following observations. First, as expected, larger $\ssub{p}{\mathrm{pos}}$ causes more performance degradation on all the models. Second, with fixed $\ssub{p}{\mathrm{pos}}$, the \nas models suffer more significant accuracy drop. For instance, on \cifarn, with $\ssub{p}{\mathrm{pos}}$ fixed as 20\%, the \cad of \nas models is 4\% higher than the manual models. Further, the \cad gap between \nas and manual models enlarges as $\ssub{p}{\mathrm{pos}}$ increases.

\begin{figure*}[!ht]
    \centering
    \includegraphics[width=170mm]{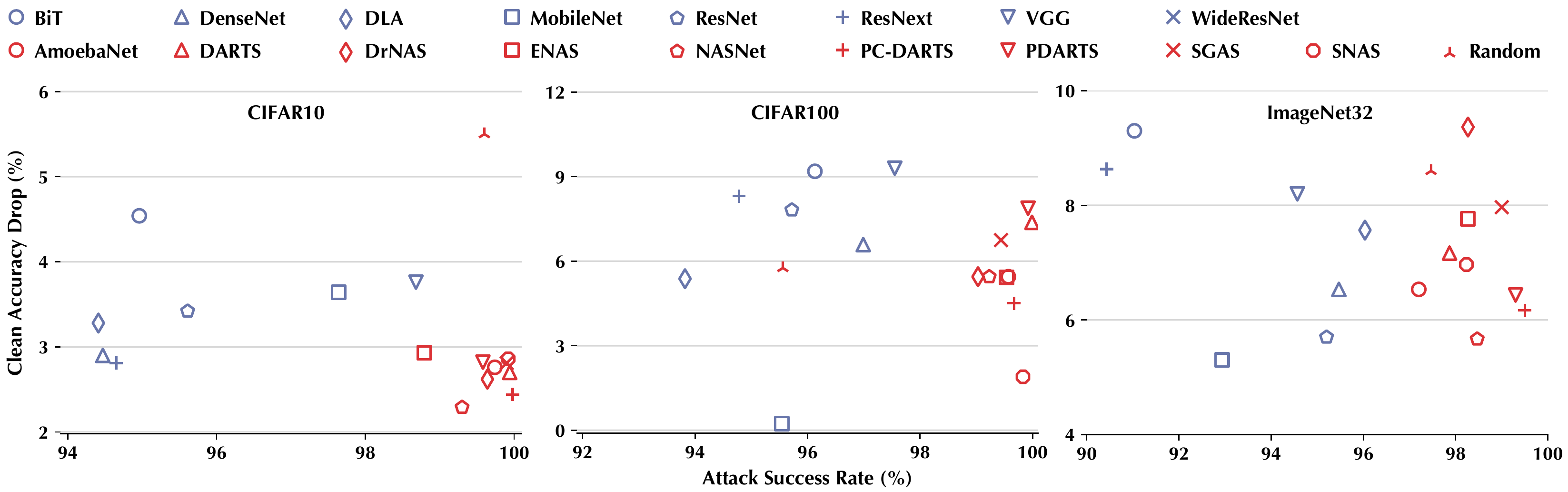}
    \caption{Performance of backdoor injection (TrojanNN) against NAS and manually designed models. \label{fig:trojannn}}
\end{figure*}

\begin{figure*}[!ht]
    \centering
    \includegraphics[width=177mm]{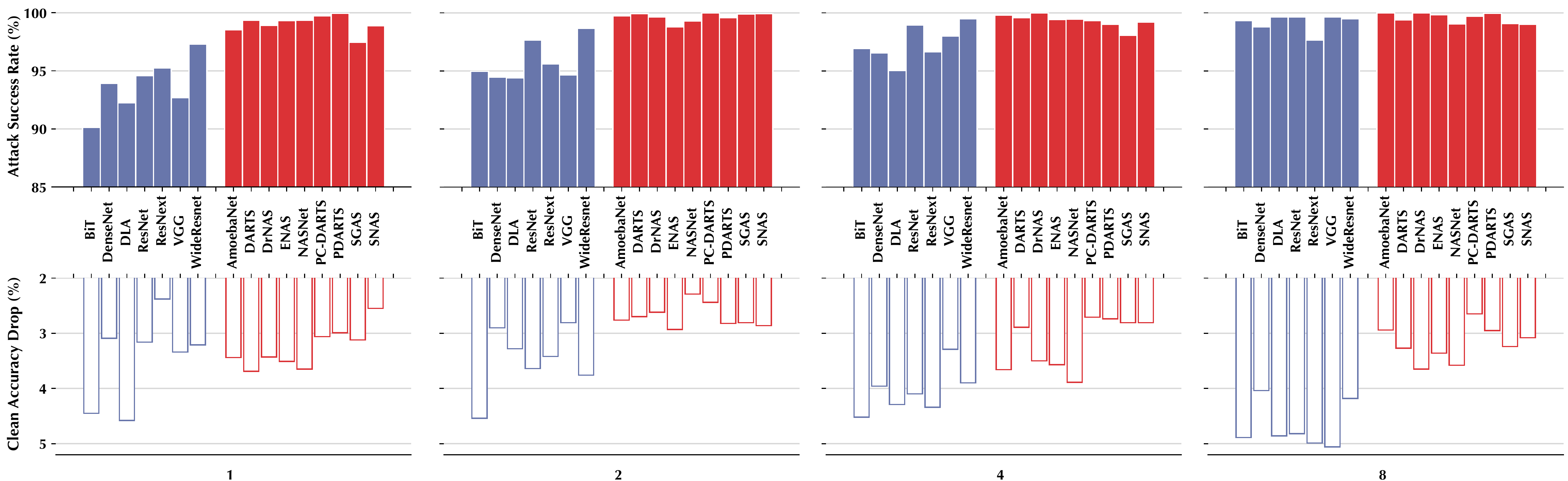}
    \caption{Impact of the number of target neurons ($n_\mathrm{neuron}$) on the vulnerability of different models with respect to \tnet on \cifar. \label{fig:neuron_num}}
\end{figure*}

\vspace{2pt}
{\bf Backdoor injection --} Next, we compare the vulnerability of \nas and manual models to neural backdoor attacks\mcite{badnet,trojannn,evil-twin}. Recall that in backdoor injection, the adversary attempts to forge a trojan model $\ssup{f}{*}$ (typically via perturbing a benign model $f$) that is sensitive to a specific trigger but behaves normally otherwise. We thus measure the attack effectiveness using two metrics: attack success rate (\asr), which is the fraction of trigger-embedded inputs successfully classified by $\ssup{f}{*}$ to the target class desired by the adversary; clean accuracy drop (\cad), which is the accuracy difference of $\ssup{f}{*}$ and $f$ on clean inputs. 

We consider {\tnet}\mcite{trojannn}, a representative backdoor attack, as the reference attack model. By optimizing both the trigger $r$ and trojan model $\ssup{f}{*}$, \tnet enhances other backdoor attacks (\meg, {\bnet}\mcite{badnet}) that employ fixed triggers. 
Figure\mref{fig:trojannn} plots the \asr and \cad of all the models, in which the results are average over 1,000 inputs randomly sampled from each testing set. Observe that the attack seems more effective against the \nas models across all the datasets. For instance, on \cifar, the attack achieves close to 100\% \asr on most \nas models with \cad below 3\%. Further, similar to adversarial evasion and model poisoning, the measures of most \nas models (except \random) are fairly consistent, indicating their similar vulnerability. Recall that \random samples models from the search space; thus, the higher vulnerability of \nas models is likely to be associated with their particular architectural properties.

We further evaluate the impact of the number of target neurons ($n_\mathrm{neuron}$) in \tnet. Recall that \tnet optimizes the trigger with respect to $n_\mathrm{neuron}$ target neurons. Figure\mref{fig:neuron_num} plots the \asr and \cad of \tnet against different models under varying setting $n_\mathrm{neuron}$. First, across all the settings of $n_\mathrm{neuron}$, \tnet consistently attains more effective attacks (\mie, higher \asr and lower \cad) on the \nas models than the manual models. Second, as $n_\mathrm{neuron}$ varies from 1 to 4, the difference of \asr between \nas and manual models decreases, while the difference of \cad tends to increase. This may be explained as follows: optimizing triggers with respect to more target neurons tends to lead to more effective attacks (\mie, higher \asr) but also result in a larger impact on clean inputs (\mie, higher \cad). However, this trade-off is less evident on the \nas models, implying their higher capabilities to fit both poisoning and clean data.

From the experiments above, we may conclude that compared with manual models, \nas models tend to be more vulnerable to backdoor injection attacks, especially under more restricted settings (\meg, fewer target neurons).

\vspace{2pt}
{\bf Functionality stealing --} We now evaluate how various models are subject to functionality stealing, in which each model $f$ as a black box only allowing query access: given input $x$, $f$ returns its prediction $f(x)$. The adversary attempts to re-construct a functionally similar model $\ssup{f}{*}$ based on the query-prediction pairs $\{(x, f(x))\}$.

\begin{figure*}[!ht]
    \centering
    \includegraphics[width=170mm]{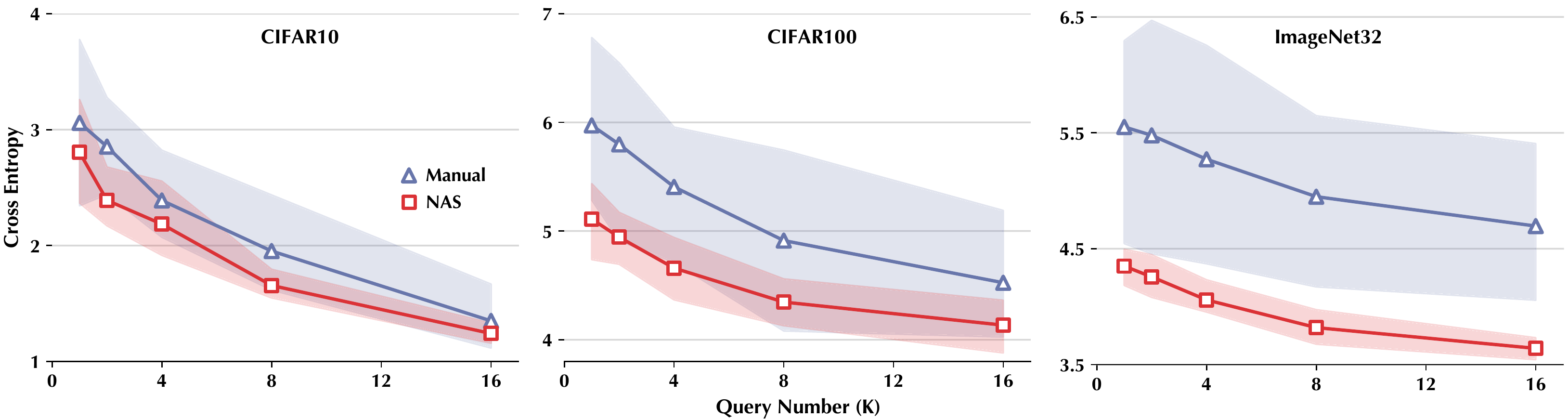}
    \caption{Performance of functionality stealing against NAS and manually designed models under the victim architecture-aware setting. \label{fig:extraction}}
\end{figure*}

\begin{figure*}[!ht]
    \centering
    \includegraphics[width=172mm]{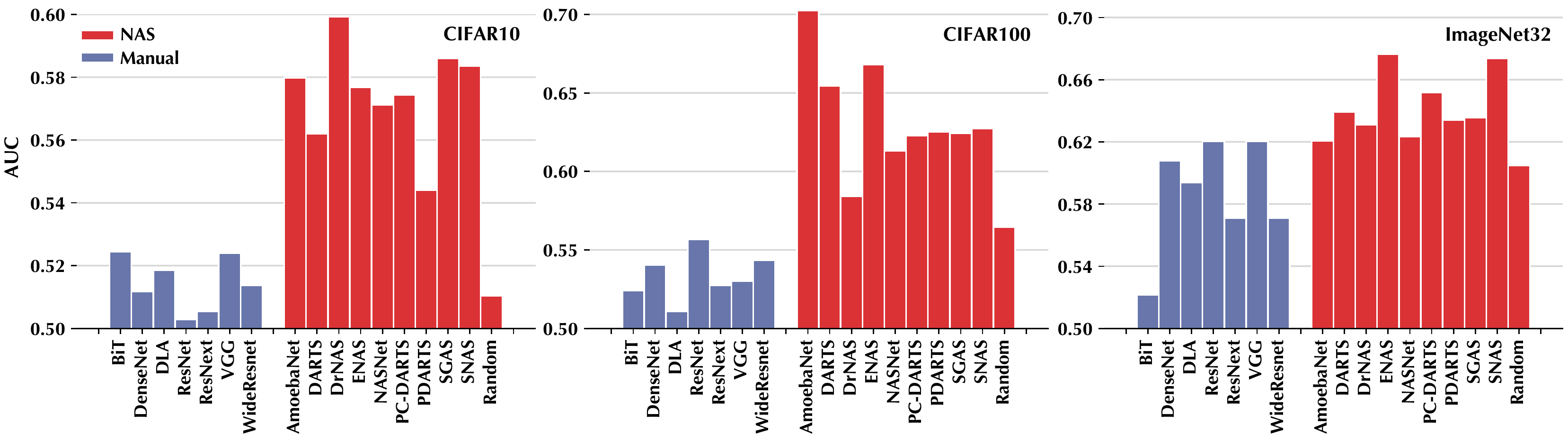}
    \caption{Performance of label-only membership inference attacks against NAS and manually designed models.  \label{fig:membership}}
\end{figure*}

We consider two scenarios: \mct{i} $f$ and $\ssup{f}{*}$ share the same architecture; and \mct{ii} the adversary is unaware of $f$'s architecture and instead uses a surrogate architecture in $\ssup{f}{*}$. We apply Knockoff\mcite{knockoff-net}, a representative functionality stealing attack that adaptively generates queries to probe $f$ to re-construct $\ssup{f}{*}$. We evaluate the attack using the average cross entropy (ACE) of $f$'s and $\ssup{f}{*}$'s predictions on the testing set, with lower cross entropy indicating more effective stealing.

\begin{table}[!ht]{\footnotesize
\centering
\renewcommand{\arraystretch}{1.2}
\setlength{\tabcolsep}{3.5pt}
\begin{tabular}{m{1.1em}c|cc|cc}
\multicolumn{2}{c|}{\multirow{2}{*}{\backslashbox{Victim $f$}{Replicate $f^*$}}} & \multicolumn{2}{c|}{Manual} & \multicolumn{2}{c}{NAS}\\
& &  {\sl ResNet} &  {\sl DenseNet} &  {\sl DARTS} &  {\sl ENAS} \\
\hline
 \multirow{2}{*}{Manual}  & {\sl ResNet} & \cellcolor[HTML]{E0DBD8}1.286 & \cellcolor[HTML]{B40426}1.509 & \cellcolor[HTML]{F7A889}1.377 & \cellcolor[HTML]{DC5D4A}1.455 \\
& {\sl DenseNet} & \cellcolor[HTML]{E1DAD6}1.288 & \cellcolor[HTML]{C7D7F0}1.245 & \cellcolor[HTML]{BED2F6}1.231 & \cellcolor[HTML]{F6A586}1.381 \\
    \hline
   \multirow{2}{*}{NAS} & {\sl DARTS} & \cellcolor[HTML]{D8DCE2}1.272 & \cellcolor[HTML]{6788EE}1.115 & \cellcolor[HTML]{93B5FE}1.172 & \cellcolor[HTML]{6E90F2}1.125 \\
  &  {\sl ENAS} & \cellcolor[HTML]{D1DAE9}1.259 & \cellcolor[HTML]{3B4CC0}1.050 & \cellcolor[HTML]{6788EE}1.115 & \cellcolor[HTML]{82A6FB}1.151 \\
  
 \end{tabular}
 \caption{Performance of functionality stealing against NAS and manual models under the victim architecture-agnostic setting. \label{tab:extraction_new}}
 }
\end{table}

Figure\mref{fig:extraction} summarizes the attack effectiveness under the victim architecture-aware setting. Across all the datasets, the attack achieves smaller ACE on the \nas models with much lower variance, in comparison with the manual models. This implies that most \nas models share similar vulnerability to functionality stealing. We further consider the victim architecture-agnostic setting. For each pair of models, we assume one as $f$ and the other as $\ssup{f}{*}$, and measure the attack effectiveness. The results on \cifar (with the query number fixed as 8K) are summarized in Table\mref{tab:extraction_new}. Observe that with the replicate model $\ssup{f}{*}$ fixed, the \nas models as the victim model $f$ result in lower ACE, implying that it tends to be easier to steal the functionality of \nas models, regardless of the architecture of the replicate model.

\vspace{2pt}
{\bf Membership inference --} Recall that in membership inference, the adversary attempts to infer whether the given input $x$ appears in the training set of the target model $f$. The inference accuracy serves as an indicator of $f$'s privacy leakages. Next, we conduct membership inference attacks on various models to assess their privacy risks. 

There are two possible scenarios: \mct{i} $f$'s prediction $f(x)$ contains the confidence score $\ssub{f}{c}(x)$ of each class $c$; and \mct{ii} $f(x)$ contains only the label $\ssup{c}{*} = \arg\max_c \ssub{f}{c}(x)$. As \mct{i} can be mitigated by removing the confidence scores in $f(x)$\mcite{membership-attack}, here, we focus on \mct{ii}. Under the class-only setting, we apply the decision boundary-based attack\mcite{label-only-membership}, which determines $x$'s membership (in the training data) by estimating its distance to the nearest decision boundary using label-only adversarial attacks (\meg, HopSkipJump\mcite{hopskipjump}). In each case, we evaluate the attack over 2,000 inputs, half randomly sampled from the training set and the other half from the testing set, and measure the attack effectiveness using the area under the ROC curve (\auc), with the estimated distance as the control of false and true positive rates. 

Figure\mref{fig:membership} compares the attack performance against different models. Notably, the attack achieves higher \auc scores on the \nas models. For instance, the average scores on the \nas and manual models on \cifarn differ by more than 0.05, while the scores on the manual models are close to random guesses (\mie, 0.5). Moreover, most \nas models (except \random) show similar vulnerability. Also, note that the manual models seem more vulnerable on \imgnet, which may be explained as follows: compared with \cifar and \cifarn, \imgnet is a more challenging dataset (see Table\mref{tab:dataset}); the models thus tend to overfit the training set more aggressively, resulting in their higher vulnerability to membership inference.
\begin{mtbox}{\small Remark\,1}
{\em \small Compared with their manually designed counterparts, \nas-generated models tend to be more vulnerable to various malicious manipulations.}
\end{mtbox}

\section{Analysis}
\label{sec:root}

The empirical evaluation in \msec{sec:measure} reveals that compared with manually designed models, \nas-generated models tend to be more vulnerable to a variety of attacks. Next, we provide possible explanations for such phenomena.

\subsection{Architectural Properties of Trainability}

We hypothesize that the greater vulnerability of \nas models stems from their key design choices. 

Popular \nas methods often evaluate the performance of a candidate model prematurely before its full convergence during the search. For instance, {\darts}\mcite{darts} formulate the search as a bi-level optimization problem, in which the inner objective optimizes a given model; to save the computational cost, instead of solving this objective exactly, it approximates the solution using a single training step, which is far from its full convergence. Similar techniques are applied in other popular \nas methods (\meg,\mcite{amoebanet,enas}). As the candidate models are not evaluated on their performance at convergence, \nas tends to favor models with higher ``trainability'' -- those converge faster during early stages -- which
result in candidate models demonstrating the following key properties:

\begin{figure*}[!ht]
    \centering
    \includegraphics[width=176mm]{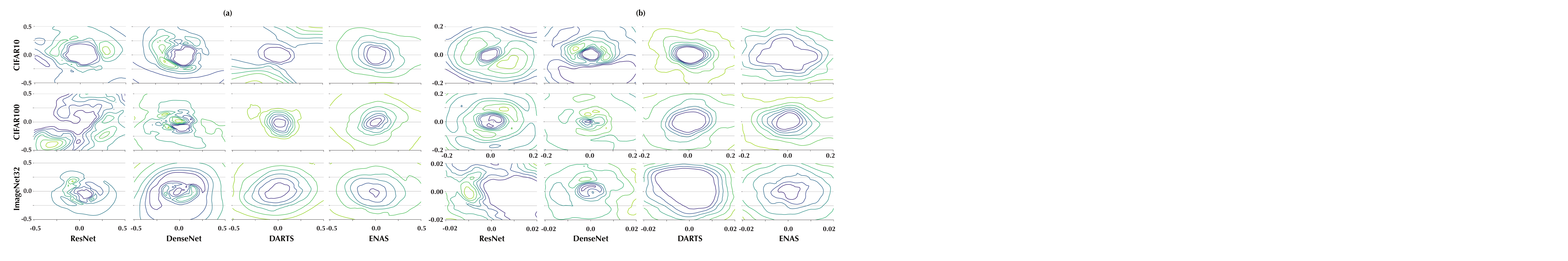}
    \caption{Loss contours of NAS-generated models (\darts, \enas) and manually designed ones (\resnet, \densenet) in (a) parameter space and (b) input space. \label{fig:loss_param}}
\end{figure*}

\vspace{2pt}
\begin{mtbox}{\small High loss smoothness}
{\em \small The loss landscape of \nas models tends to be smooth, while the gradient provides effective guidance for optimization. Therefore, \nas models are amenable to training using simple, first-order optimizers.}
\end{mtbox}

\vspace{2pt}
\begin{mtbox}{\small Low gradient variance}
{\em \small The gradient of \nas models with respect to the given distribution tends to have low variance. Therefore, the stochastic gradient serves as a reliable estimate of the true gradient, making \nas models converge fast.}
\end{mtbox}

Note that the loss smoothness captures the geometry of the loss function in the parameter space (or the input space), while the gradient variance measures the difference between the gradients with respect to different inputs. While related, the former dictates whether a model is easy to train if the gradient direction is known and the latter dictates whether it is easy to estimate the gradient direction reliably.

\begin{figure*}[!ht]
    \centering
    \includegraphics[width=170mm]{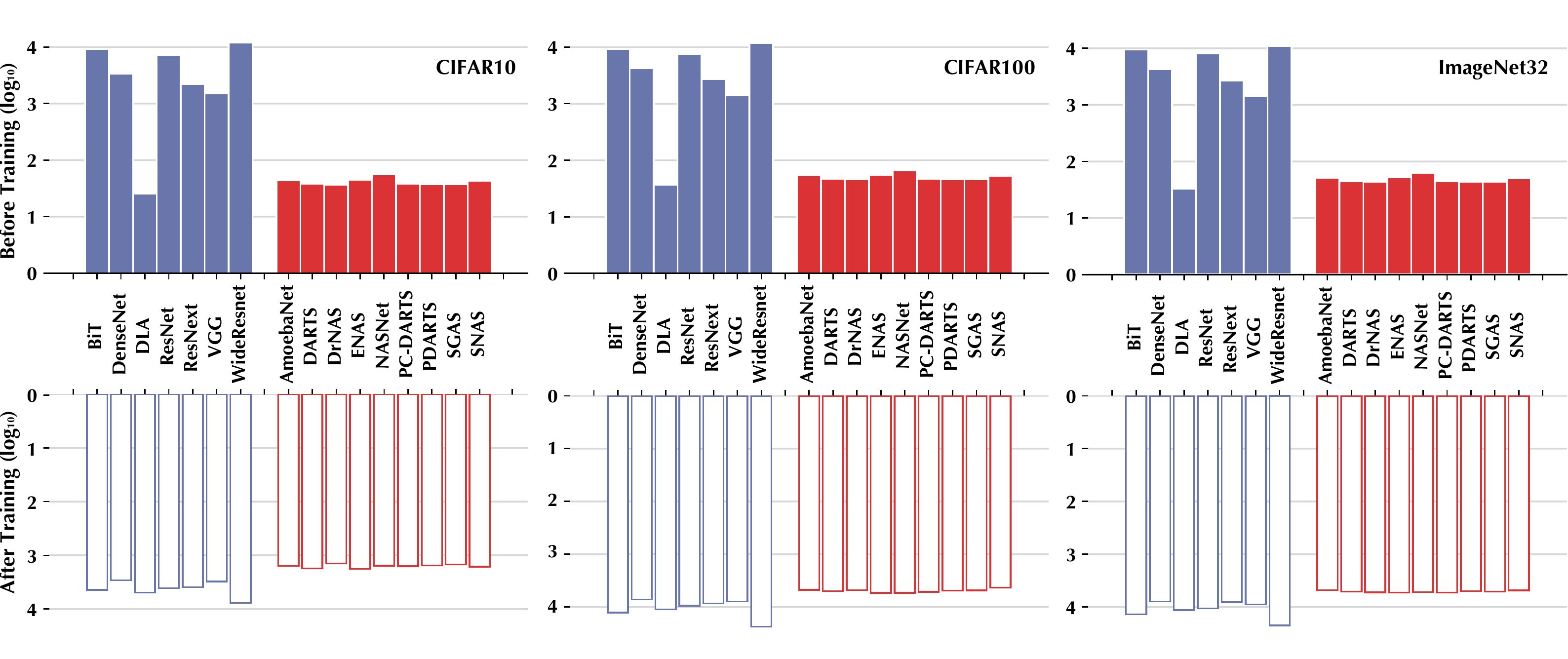}
    \caption{Gradient variance of NAS-generated and manually designed models before and after training. \label{fig:grad_var}}
\end{figure*}

Next, we empirically validate the above hypotheses by comparing the gradient smoothness and variance of \nas-generated and manually designed models.

\vspace{2pt}
{\bf Loss smoothness --} A loss function $\gL$ is said to have $L$-Lipschitz ($L > 0$) continuous gradient with respect to $\theta$ if it satisfies $\| \nabla \gL(\theta) -\nabla \gL(\theta')\|\leq L \| \theta - \theta'\|$ for any $\theta, \theta'$. The  constant $L$ controls $\gL$'s smoothness. While it is difficult to directly measure $L$ of given model $f$, we explore its loss contour\mcite{loss-landscape}, which quantifies the impact of parameter perturbation on $\gL$. Specifically, we measure the loss contour of model $f$ as follows:
\begin{equation}
\Gamma(\alpha, \beta) = \gL(\ssup{\theta}{*} + \alpha \ssub{d}{1} + \beta \ssub{d}{2})
\end{equation}
where $\ssup{\theta}{*}$ denotes the local optimum, $\ssub{d}{1}$ and $\ssub{d}{2}$ are two random, orthogonal directions as the axes, and $\alpha$ and $\beta$ represent the perturbation steps along $\ssub{d}{1}$ and $\ssub{d}{1}$, respectively. Notably, the loss contour effectively approximates the loss landscape in a two-dimensional space\mcite{loss-vis}. 

Figure\mref{fig:loss_param}(a) visualizes the loss contours of \nas (\darts and \enas) and manual (\resnet and \densenet) models across different datasets. Observe that the \nas models tend to demonstrate a flatter loss landscape. Similar phenomena are observed with respect to other models. This observation may explain why the gradient of \nas models gives more effective guidance for minimizing the loss function, leading to their higher trainability. 

Further, for the purpose of the analysis in \msec{sec:root}, we extend the loss smoothness in the parameter space to the input space. We have the following result to show their fundamental connections (proof deferred to \msec{sec:proof}). 

\begin{theorem}
\label{the:smoothness}
If the loss function $\gL$ has $L$-Lipschitz continuous gradient with respect to $\theta$ and the weight matrix of each layer of the model is normalized\mcite{weight-normalization}, then $\gL$ has $L/\sqrt{n}$-Lipschitz continuous gradient with respect to the input, where $n$ is the input dimensionality.
\end{theorem}

Empirically, we define $f$'s loss contour with respect to a given input-class pair $(x, y)$ as follows:
\begin{equation}
\ssub{\Gamma}{(x,y)}(\alpha, \beta) = \ell(f(x + \alpha \ssub{d}{1} + \beta \ssub{d}{2}), y)
\end{equation}
where $\ssub{d}{1}$ and $\ssub{d}{2}$ are two random, orthogonal directions in the input space.
Figure\mref{fig:loss_param}(b) visualizes the loss contours of \nas and manual models in the vicinity of randomly sampled inputs. It is observed that \nas models also demonstrate higher loss smoothness in the input space, compared with the manual models.

\vspace{2pt}
{\bf Gradient variance --} Meanwhile, the variance of the gradient with respect to inputs sampled from the underlying  distribution quantifies the noise level of the gradient estimate used by stochastic training methods (\meg, SGD)\mcite{gradient-deviation}. Formally, let $g$ be the stochastic gradient. We define the gradient variance as follows (where the expectation is taken with respect to the given distribution):
\begin{equation}
\text{\small Var}(g) = \sE \left[ \| g  -  \sE\left [ g \right]\|_2^2 \right]
\end{equation}

Assuming $g$ is an unbiased estimate of the true gradient,  $\text{\small Var}(g)$ measures $g$'s expected deviation from the true gradient. Smaller $\text{\small Var}(g)$ implies lower noise level, thereby more stable updating of the model parameters $\theta$.

In Figure\mref{fig:grad_var}, we measure the gradient variance of various models before training (with Kaiming initialization\mcite{kaiming-init}) and after training is complete. It is observed in all the cases that at initialization, the gradient variance of \nas models is more than two orders of magnitude smaller than the manual models and then grows gradually during the training; in comparison, the gradient variance of manual models does not change significantly before and after training. This observation may explain why the stochastic gradient of \nas models gives a reliable estimate of the true gradient, making them converge fast at early training phases.

\subsection{Explanations of Attack Vulnerability}

We now discuss how the vulnerability of \nas models to various attacks can be attributed to the properties of high loss smoothness and low gradient variance.

\vspace{2pt}
{\bf Adversarial evasion --} The vulnerability to adversarial evasion is mainly attributed to the sensitivity of model prediction $f(x)$ to the perturbation of input $x$. Under the white-box setting, the adversary typically relies on the gradient to craft the adversarial input $\ssup{x}{*}$. For instance, {\pgd}\mcite{pgd} crafts $\ssup{x}{*}$ by iteratively updating the input using the following rule: 
\begin{equation}
    \ssub{x}{t+1} = \ssub{\Pi}{x + \ssub{\gB}{\epsilon}} \left(\ssub{x}{t} + \alpha \sign \left(\ssub{\nabla}{x}\, \ell (f(\ssub{x}{t}), y)\right)\right) 
\end{equation}
where $\ssub{x}{t}$ is the perturbed input after the $t$-th iteration, $\Pi$ denotes the projection operator, $\ssub{\gB}{\epsilon}$ represents the allowed set of perturbation (parameterized by $\epsilon$), and $\alpha$ is the perturbation step. Apparently, the attack effectiveness relies on whether the gradient $\ssub{\nabla}{x}\, \ell (f(\ssub{x}{t}), y)$ is able to provide effective guidance for perturbing $\ssub{x}{t}$. 

As shown in \msec{sec:analysis}, compared with the manual models, due to the pursuit of higher trainability, the \nas models often demonstrate a smoother loss landscape wherein the gradient at each point represents effective optimization direction; thus, the \nas models tend to be more vulnerable to gradient-based adversarial evasion. Notably, this finding also corroborates existing studies (\meg, \mcite{goodfellow:fsgm}) on the fundamental tension between designing ``linear'' models that are easier to train and designing ``nonlinear'' models that are more resistant to adversarial evasion.

The similar phenomena observed in the case of black-box attacks (\meg, \nes) may be explained as follows: to perform effective perturbation, black-box attacks often rely on indirect gradient estimation, while the high loss smoothness and low gradient variance of \nas models lead to more accurate and efficient (with fewer queries) gradient estimation.


\vspace{2pt}
{\bf Model poisoning --} The vulnerability to model poisoning can be attributed to the sensitivity of model training to the poisoning data in the training set. Here, we analyze how the property of low gradient variance impacts this sensitivity. 

For a given dataset $\gD$, let $\gL(\theta)$ be the loss of a model $\ssub{f}{\theta}$ parameterized by $\theta$ with respect to $\gD$:
\begin{equation}
\gL(\theta) \triangleq \frac{1}{|\gD|}\ssub{\sum}{(x, y) \in \gD} \ell(\ssub{f}{\theta}(x), y)
\end{equation} 
Further, let $\ssup{\theta}{*}$ represent $f$'s (local) optimum with respect to $\gD$.
With $\theta$ initialized as $\ssub{\theta}{0}$, consider $T$-step SGD updates with the $t$-th step update as:
\begin{equation}
    \ssub{\theta}{t+1} = \ssub{\theta}{t} -\ssub{\alpha}{t} \ssub{g}{t}
\end{equation}
where $\ssub{\alpha}{t}$ is the step size and $\ssub{g}{t}$ is the gradient estimate. 
We have the following result describing the convergence property of $\ssub{\theta}{t}$ $(t = 1, \ldots, T)$.
\begin{theorem}[\mcite{gradient-deviation}] 
\label{the:converge}
 Assuming that (i) $\gL(\theta)$ is continuous and differentiable, with its gradient bounded by Lipschitz constant $L$, (ii) the variance of gradient estimate $\ssub{g}{t}$ $(t = 1, \ldots, T)$ is bounded by $\ssup{\sigma}{2}$, and (iii) $\ssub{\theta}{t}$ is selected as the final parameters with probability proportional to $2 \ssub{\alpha}{t}  - L\sboth{\alpha}{t}{2}$. Then, the output parameters $\ssub{\theta}{\bar{t}}$ satisfies:
\begin{equation}
    \sE\left[\gL (\ssub{\theta}{\bar{t}}) - \gL (\ssup{\theta}{*})\right] \leq \frac{ \| \ssub{\theta}{0} - \ssup{\theta}{*} \|^2 + \ssup{\sigma}{2} \sboth{\sum}{t=1}{T}\sboth{\alpha}{t}{2}   }{\sboth{\sum}{t=1}{T} (2 \ssub{\alpha}{t}  - L\sboth{\alpha}{t}{2})}
\end{equation}
where the expectation is defined with respect to the selection of $\bar{t}$ and the gradient variance.
\end{theorem}

Intuitively, Theorem\mref{the:converge} describes the properties that impact the fitting of model $f$ to the given dataset $\gD$. As shown in \msec{sec:analysis}, compared with the manual models, the \nas models tend to have both higher loss smoothness (\mie, smaller $L$) and lower gradient variance (\mie, smaller $\sigma$). Therefore, the \nas models tend to fit $\gD$ more easily. Recall that in model poisoning, $\gD$ consists of both clean data $\trnd$ and poisoning data $\posd$,  fitting to $\gD$ more tightly implies more performance drop over the testing data, which may explain the greater vulnerability of \nas models to model poisoning. 

\vspace{2pt}
{\bf Backdoor injection --} Recall that in backdoor injection, the adversary forges a trojan model $\ssup{f}{*}$ that is sensitive to a trigger pattern $r$ such that any input $x$, once embedded with $r$, tends to be misclassified to a target class $t$: $\ssup{f}{*}(x+r) = t$. To train $\ssup{f}{*}$, the adversary typically pollutes the training data $\trnd$ with trigger-embedded inputs. 

Intuitively, this attack essentially exploits the attack vectors of adversarial evasion that perturbs $x$ at inference time and model poisoning that pollutes $\trnd$ at training time. Therefore, the vulnerability of \nas models to both attack vectors naturally results in their vulnerability to backdoor injection. Due to the space limitations, we omit the detailed analysis here.

\vspace{2pt}
{\bf Functionality stealing --} Recall that in functionality stealing (\meg, Knockoff\mcite{knockoff-net}), the adversary (adaptively) generates queries to probe the victim model $f$ to replicate a functionally similar one $\ssup{f}{*}$. For instance, Knockoff encourages queries that are certain by $f$, diverse across different classes, and disagreed by $\ssup{f}{*}$ and $f$. 

The effectiveness of such attacks depends on $f$'s loss landscape with respect to the underlying distribution; intuitively, the complexity of the loss landscape in the input space implies the hardness of fitting $\ssup{f}{*}$ to $f$ based on a limited number of queries. Thus, given their high loss smoothness, the \nas models tend to be more vulnerable to functionality stealing.

\begin{figure}[!ht]
    \centering
    \includegraphics[width=50mm]{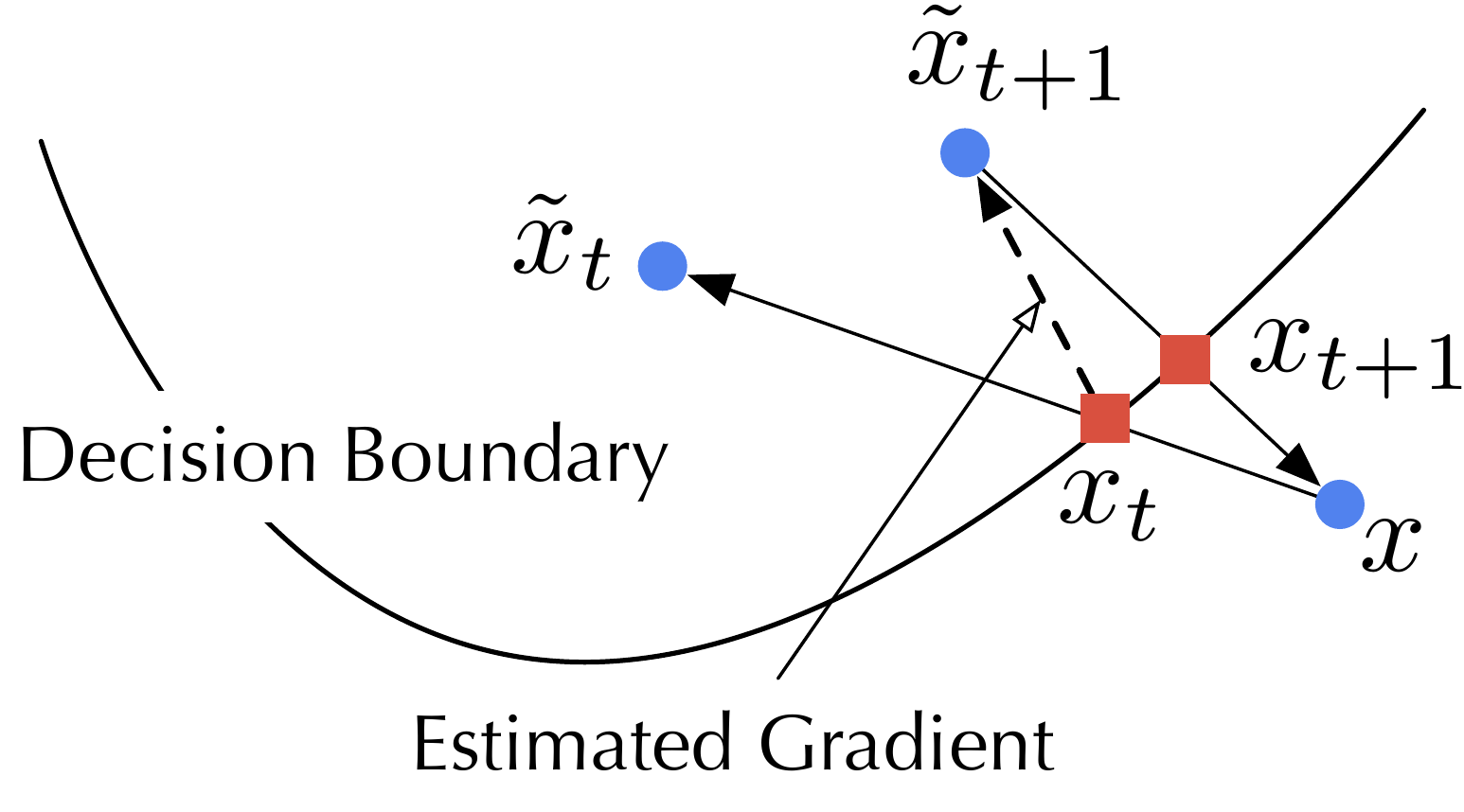}
    \caption{Illustration of the HopSkipJump attack. \label{fig:hsj}}
\end{figure}

\vspace{2pt}
{\bf Membership inference --} 
It is shown in \msec{sec:measure} that the \nas models seem more vulnerable to membership inference, especially under the label-only setting in which only the prediction labels are accessible. The adversary thus relies on signals such as input $x$'s distance to its nearest decision boundary $\text{dist}(x, f(x))$; intuitively, if $x$ appears in the training set, $\text{dist}(x, f(x))$ is likely to be below a certain threshold. Concretely,
the \hsj attack\mcite{hopskipjump} is employed in\mcite{label-only-membership} to estimate 
$\text{dist}(x, f(x))$ via iteratively querying $f$ to find point $\ssub{x}{t}$ on the decision boundary using bin search, walking along the boundary using the estimated gradient at $\ssub{x}{t}$, and finding point $\ssub{x}{t+1}$ to further reduce the distance to $x$, which is illustrated in Figure\mref{fig:hsj}.

The effectiveness of this attack hinges on \mct{i} the quality of estimated gradient and \mct{ii} the feasibility of descending along the decision boundary. For the \nas models, the gradient estimate tends to be more accurate due to the low gradient variance, while the decision boundary tends to be smoother due to the high loss smoothness, which may explain the greater vulnerability of \nas models to label-only membership inference attacks.

\begin{mtbox}{\small Remark\,2}
{\em \small The high loss smoothness and low gradient variance of \nas-generated models may account for their greater vulnerability to various attacks.}
\end{mtbox}

\subsection{Connections of Various Attacks} 

It is shown above that the vulnerability of \nas models to various attacks may be explained by their high loss smoothness and low gradient variance, which bears an intriguing implication: different attacks may also be inherently connected via these two factors.

Specifically, most existing attacks involve input or model perturbation. For instance, adversarial evasion, regardless of the white- or black-box setting, iteratively computes (or estimates) the gradient and performs perturbation accordingly;  backdoor injection optimizes the trigger and model jointly, requiring to estimate, based on the gradient, how the model responds to the updated trigger. 

The effectiveness of such attacks thus highly depends on \mct{i} how to estimate the gradient at each iteration and \mct{ii} how to use the gradient estimate to guide the input or model perturbation. Interestingly, gradient variance and loss smoothness greatly impact \mct{i} and \mct{ii}, respectively: low gradient variance enables the adversary to accurately estimate the gradient, while high loss smoothness allows the adversary to use such estimate to perform effective perturbation.
\begin{mtbox}{\small Remark\,3}
{\em \small The effectiveness of various attacks is inherently connected through loss smoothness and gradient variance.}
\end{mtbox}

\section{Discussion}
\label{sec:discussion} 

In \msec{sec:measure} and \msec{sec:root}, we reveal the relationships between the trainability of \nas-generated models and their vulnerability to various attacks, two key questions remain: \mct{i} what are the architectural patterns associated with such vulnerability? and \mct{iii} what are the potential strategies to remedy the vulnerability incurred by the current \nas practice? In this section, we explore these two questions and further discuss the limitations of this work.

\subsection{Architectural Weaknesses} 
\label{sec:weakness}

As shown in \msec{sec:root}, the vulnerability of \nas models is potentially related to their high loss smoothness and low gradient variance, which stem from the preference for models of higher trainability. We now discuss how such preference is reflected in the concrete architectural patterns, which we examine from two aspects, namely, topology selection and operation selection.

\begin{table}[!ht]{\footnotesize
\centering
\renewcommand{\arraystretch}{1.2}
\begin{tabular}{r|ccc}
 Architecture & Cell Depth & Cell Width & \# Skip connects \\
  \hline
  {\sl AmoebaNet} & 4 & 3$c$ & 2\\
  {\sl DARTS} &  3 & 3$c$ & 3\\
 {\sl DrNAS} & 4 & 2$c$ & 1\\
  {\sl ENAS}  &  2 & 5$c$ & 2 \\
  {\sl NASNet} & 2 & 5$c$ & 1 \\
  {\sl PC-DARTS} & 2 & 4$c$ & 1 \\
  {\sl PDARTS} & 4 & 2$c$ & 2  \\
 {\sl SGAS} & 3 & 2$c$ & 1 \\
  {\sl SNAS} & 2 & 4$c$ & 4\\
 \end{tabular}
 \caption{The cell depth and width, and the number of skip connects of representative NAS-generated models (the width of each intermediate node is assumed to be $c$). \label{tab:depth-width}}
 }
\end{table}

\vspace{2pt}
{\bf Topology selection --} Recent studies\mcite{shallow-wide-cell} suggest that in cell-based \nas, the preference for models with faster convergence often results in wide, shallow cell structures. As shown in Figure\mref{fig:automl}, the cell depth is defined as the number of connections along the longest path from the input nodes to the output node; the width of each intermediate node is defined as the number of channels for convolution operators or the number of features for linear operators, while the cell width is defined as the total width of intermediate nodes connected to the input nodes. Table\mref{tab:depth-width} summarizes the cell depth and width of \nas models used in our evaluation. It is observed that the cell structures of most \nas models are both shallow (with an average depth of 2.8) and wide (with an average width of 3.3$c$), where the width of each intermediate node is assumed to be $c$.

It is shown in\mcite{shallow-wide-cell} that under similar settings (\mie, the same number of nodes and connections), wide and shallow cells tend to demonstrate higher trainability. This observation is also corroborated by the recent theoretical studies on the convergence of wide neural networks\mcite{wide-dnn}: neural networks of infinite width tend to evolve as linear models using gradient descent optimization.

\vspace{2pt}
{\bf Operation selection --} The preference for higher trainability also impacts the selection of operations (\meg, 3$\times$3 convolution versus skip connection) on the connections within the cell structure, and typically favors skip connects over other operations.

Recall that differential \nas methods\mcite{darts,drnas,sgas} typically apply continuous relaxation on the search space to enable direct gradient-based optimization. The operation on each connection is modeled as a softmax of all possible operations $\gO$ and discretized by selecting the most likely one $\arg \ssub{\max}{o \in \gO} \ssub{\alpha}{o}$. It is shown in\mcite{rethinking-nas} that in well-optimized models, the weight of skip connection $\ssub{\alpha}{\text{skip}}$ often exceeds other operations, leading to its higher chance of being selected. This preference takes effect in our context, as \nas models tend to converge fast at early training stages. Table\mref{tab:depth-width} summarizes the number of skip connects in each cell of representative \nas models. Observe that most \nas models have more than one skip connection in each cell.

The operation of skip connection is originally designed to enable back-propagation in DNNs\mcite{resnet,densenet}. As a side effect, accurate gradient estimation also facilitates attacks that exploit gradient information\mcite{skip-connection}. Thus, the over-use of skip connects in \nas models also partially accounts for their vulnerability to such attacks. 

\begin{mtbox}{\small Remark\,4}
{\em \small \nas-generated models often feature wide and shallow cell structures as well as overuse of skip connects.}
\end{mtbox}

\subsection{Potential Mitigation} 
\label{sec:mitigation}

We now discuss potential mitigation to remedy the vulnerability incurred by the \nas practice. We consider enhancing the robustness of \nas models under both post-\nas and in-\nas settings. In post-\nas mitigation, we explore using existing defenses against given attacks to enhance \nas models, while with in-\nas mitigation, we explore building attack robustness into the \nas process directly. 

\vspace{2pt}
{\bf Post-NAS mitigation --} As a concrete example, we apply adversarial training\mcite{Madry:2017:iclr,adv-train-free}, one representative defense against adversarial evasion, to enhance the robustness of \nas models. Intuitively, adversarial training improves the robustness of given model $f$ by iteratively generating adversarial inputs with respect to its current configuration and updating $f$ to correctly classify such inputs.

\begin{figure}[!ht]
    \centering
    \includegraphics[width=82mm]{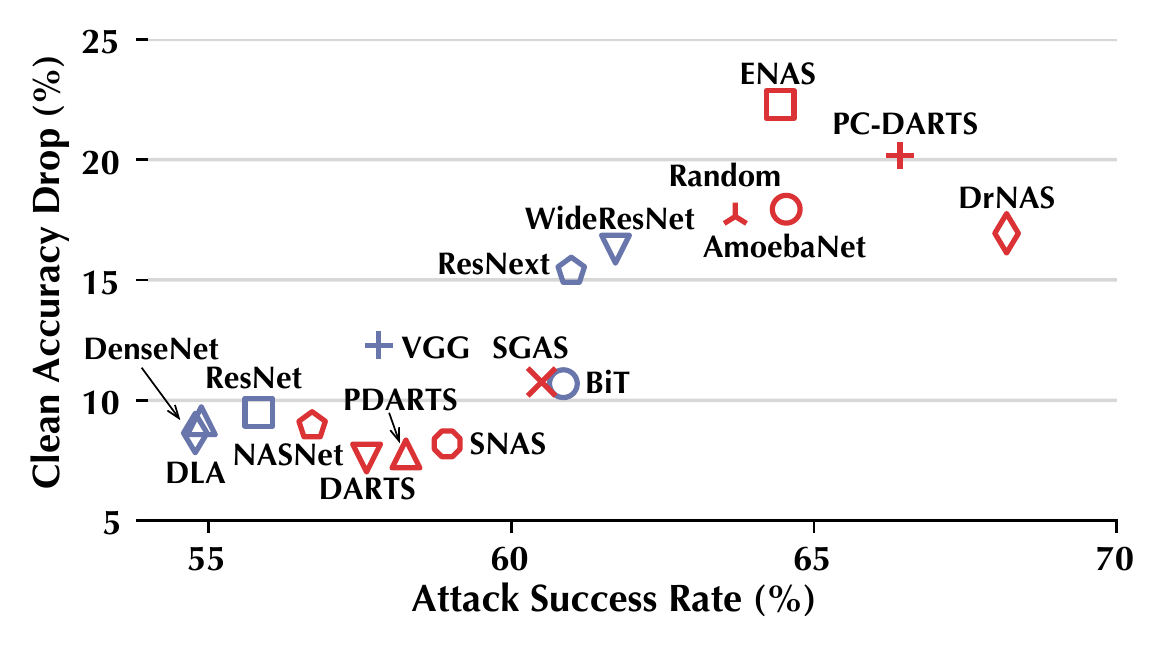}
    \caption{Effectiveness of adversarial training on various models over \cifar. \label{fig:adv_train_cifar10}}
\end{figure}

Figure\mref{fig:adv_train_cifar10} compares the effectiveness of adversarial training on various models over \cifar. For each model, we measure its accuracy (in terms of accuracy drop from before adversarial training) and robustness (in terms of the success rate of the untargeted \pgd attack). Observe that a few \nas models (\meg, \darts) show accuracy and robustness comparable with manual models, while the other \nas models (\meg, \drnas) underperform in terms of both accuracy and robustness, which may be explained by their diverse architectural patterns associated with adversarial training (\meg, dense connections, number of convolution operations, and cell sizes)\mcite{robust-nas}. This disparity also implies that adversarial training may not be a universal solution for improving the robustness of all the \nas models.

\vspace{2pt}
{\bf In-\nas mitigation --} We further investigate how to build attack robustness into the \nas process directly. Motivated by the analysis in \msec{sec:weakness}, we explore two potential strategies.

\begin{figure}[!ht]
    \centering
    \includegraphics[width=80mm]{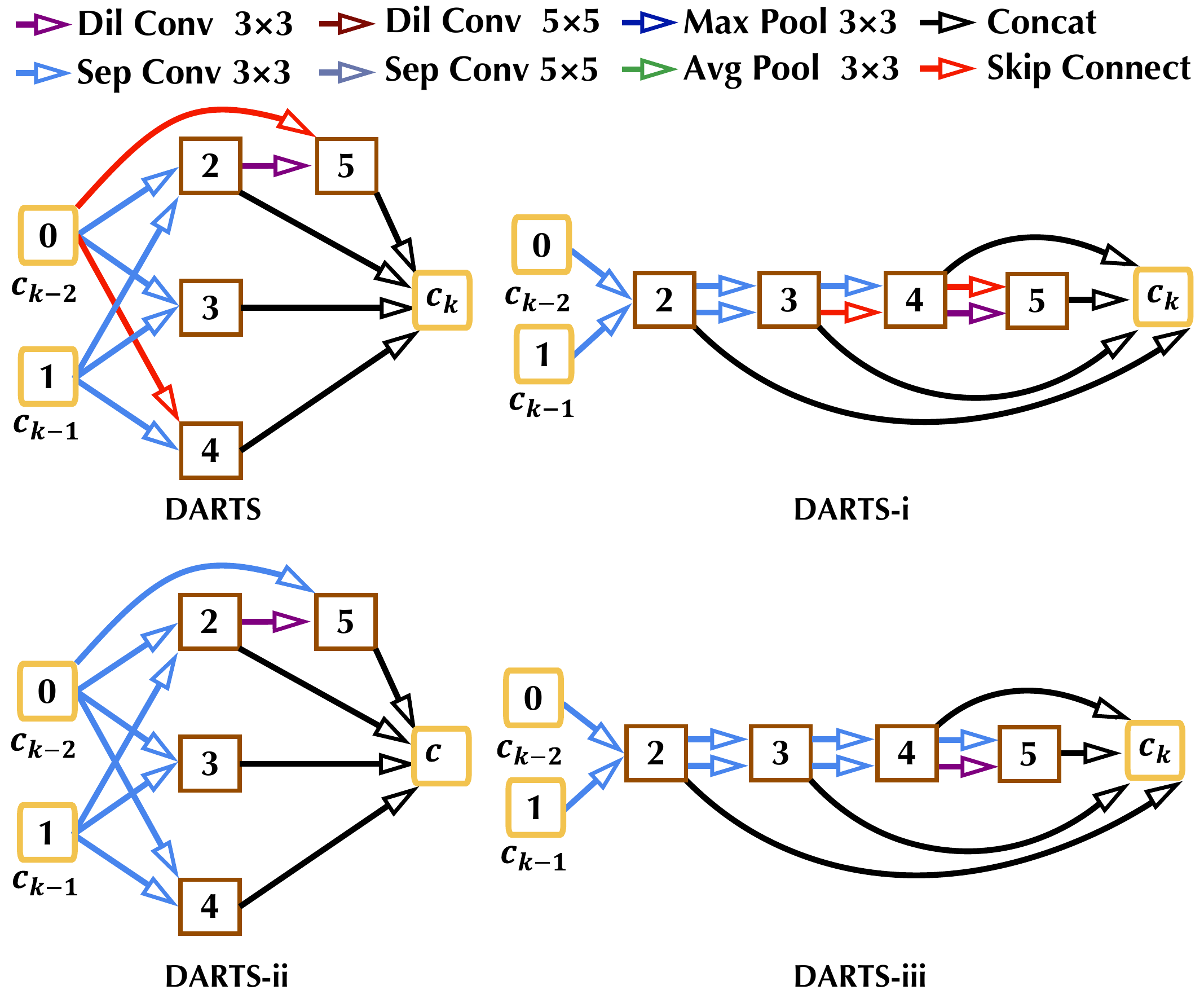}
    \caption{Illustration of cell structures of \darts, \darts-{\sl i}, \darts-{\sl ii}, and \darts-{\sl iii}. \label{fig:arch}}
\end{figure}

\mct{i} Increasing cell depth -- As the vulnerability of \nas models tends to be associated with their wide and shallow cell structures, we explore increasing their cell depth. To this end, we may re-wire existing \nas models or modify the performance measure of candidate models. For the latter case, we may increase the number of training epochs before evaluation. For instance, \darts, without fully optimizing model parameters $\theta$ with respect to architecture parameters $\alpha$, uses a single-step gradient descent ($\ssub{n}{\text{step}} = 1$) to approximate the solution\mcite{darts}. We improve the approximation by increasing the number of training steps (\meg, $\ssub{n}{\text{step}} = 5$) at the cost of additional search time.

\mct{ii} Suppressing skip connects -- As the vulnerability of \nas models is also associated with skip connects, we explore purposely reducing their overuse. To this end, we may replace the skip connects in existing \nas models with other operations (\meg, convolution) or modify their likelihood of being selected in the search process. For the latter case, at each iteration, we may multiply the weight of skip connection $\ssub{\alpha}{\text{skip}}$ by a coefficient $\gamma \in (0, 1)$ in \meq{eq:softmax}.

We evaluate the effectiveness of such strategies within the \darts framework. Let \darts-{\sl i}, \darts-{\sl ii}, and \darts-{\sl iii} be the variants of \darts after applying the strategies of \mct{i}, \mct{ii}, and \mct{i} and \mct{ii} combined. Figure\mref{fig:arch} compares their cell structures. Notably, \darts-{\sl i} features a cell structure deeper than \darts (5 versus 2), while \darts-{\sl ii} and \darts-{\sl iii} substitute the skip connects in \darts and \darts-{\sl i} with $3\times3$ convolution, respectively.

\begin{table}[!ht]{\footnotesize
\centering
\renewcommand{\arraystretch}{1.2}
\setlength{\tabcolsep}{4pt}
\begin{tabular}{l|cc|cc|c}
\multirow{2}{*}{Architecture} & \multicolumn{2}{c|}{Evasion} & \multicolumn{2}{c|}{Backdoor} & Membership  \\
\cline{2-6}
& ASR (M) & ASR (L) & ASR & CAD & AUC \\
\hline
\hline
{\sl DARTS} & 100.0\% & 86.7\% & 99.9\% & 2.7\% & 0.562 \\
{\sl DARTS-i} & 88.3\% & 72.7\% & 90.4\% & 4.6\% & 0.534 \\
{\sl DARTS-ii} & 93.0\% & 75.0\% & 98.8\% & 3.0\% & 0.531 \\
{\sl DARTS-iii} & 82.0\% & 65.6\% & 84.2\% & 4.6\% & 0.527 \\
 \end{tabular}
 \caption{Vulnerability of {\sl DARTS} and its variants to adversarial evasion (M - most likely case, L - least likely case), backdoor injection, and membership inference on \cifar.  \label{tab:mitigation}}
 }
\end{table}

\begin{figure}[!ht]
    \centering
    \includegraphics[width=75mm]{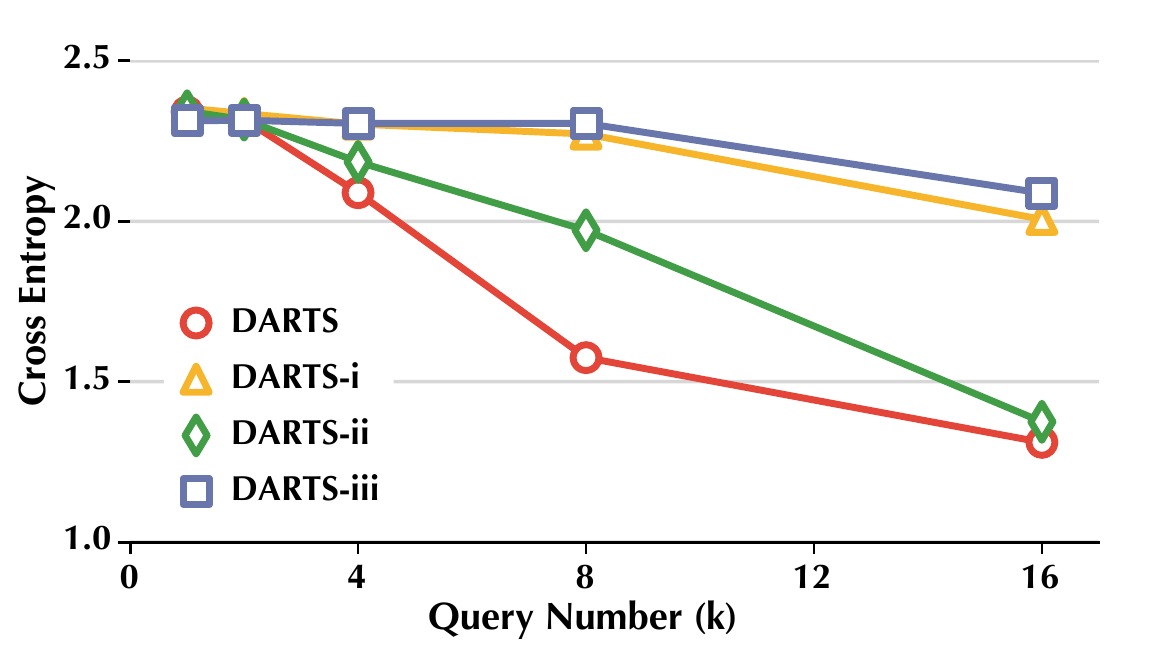}
    \caption{Vulnerability of {\sl DARTS} and its variants to model extraction on \cifar. \label{fig:mitigation_extraction_loss}}
\end{figure}

Table\mref{tab:mitigation} compares their vulnerability to adversarial evasion, backdoor injection, and membership inference on \cifar. The experimental setting is identical to that in \msec{sec:measure}. Observe that both strategies may improve the robustness of \nas models against these attacks. For instance, combining both strategies in \darts-{\sl iii} reduces the \auc score of membership inference from 0.562 to 0.527. Similar phenomena are observed in the case of model extraction attacks. As shown in Figure\mref{fig:mitigation_extraction_loss}, increasing the cell depth significantly augments the robustness against model extraction, while suppressing skip connects further improves it marginally.

\begin{figure}[!ht]
    \centering
    \includegraphics[width=75mm]{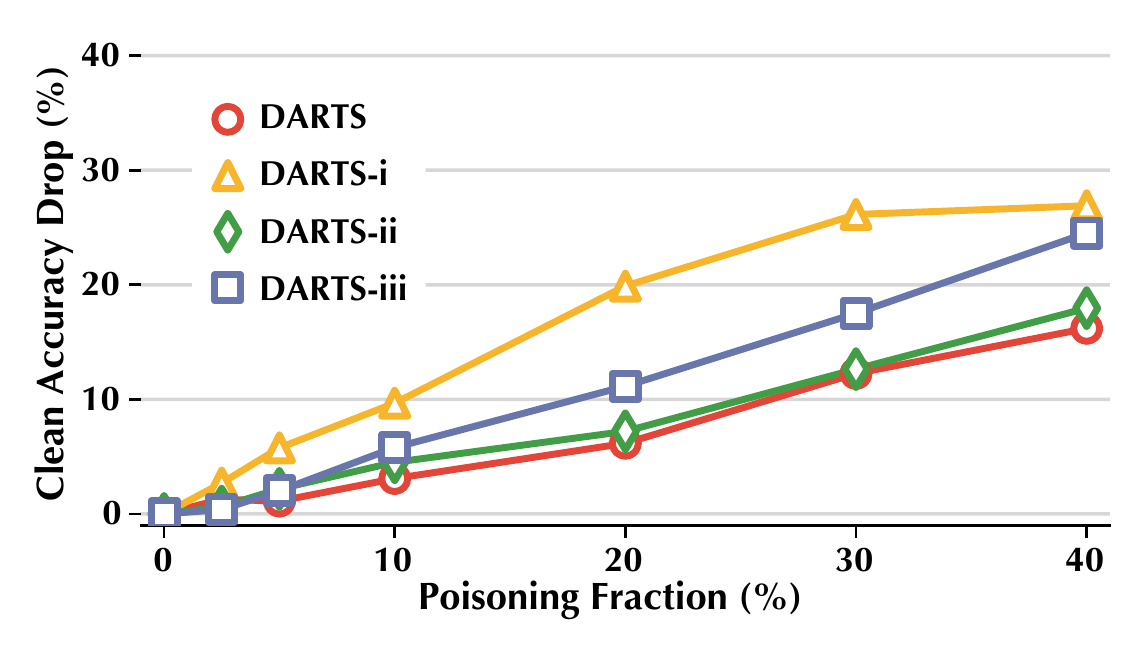}
    \caption{Vulnerability of {\sl DARTS} and its variants to model poisoning on \cifar. \label{fig:mitigation_poison}}
\end{figure}

Yet, such strategies seem to have a negative impact on the robustness against model poisoning. As shown in Figure\mref{fig:mitigation_poison}, both strategies, especially increasing the cell depth, tends to exacerbate the attack vulnerability. This may be explained by that while more difficult to fit the poisoning data, it is also more difficult to fit deeper structures to the clean data, which results in a significant accuracy drop. This may also explain why the backdoor injection attack has higher \cad on \darts-{\sl i} and \darts-{\sl iii} as shown in Table\mref{tab:mitigation}. The observation also implies a potential trade-off between the robustness against different attacks in designing \nas models.

\begin{mtbox}{\small Remark\,5}
{\em \small Simply increasing cell depth and/or suppressing skip connects may only partially mitigate the vulnerability of \nas-generated models.}
\end{mtbox}

\subsection{Limitations}

Next, we discuss the limitations of this work. 

\vspace{2pt}
{\bf Alternative \nas frameworks --} In this work, we mainly consider the cell-based search space adopted by recent NAS methods\mcite{enas,darts,darts-r,darts-,drnas}, while other methods have considered the global search space (\meg, chain-of-layer structures)\mcite{rl-automl,path-level}. Further, while we focus on the differentiable search strategy, there are other strategies including random search\mcite{random-perm}, Bayesian optimization\mcite{bayesian-nas}, and reinforcement learning\mcite{rl-automl,nasnet,rl-nas-c}. We consider exploring the vulnerability of models generated by alternative \nas frameworks as our ongoing research.

\vspace{2pt}
{\bf Other trainability metrics --} In this work, we only consider loss smoothness and gradient variance as two key factors impacting the trainability (and vulnerability) of \nas models. There are other trainability metrics (\meg, condition number of neural tangent kernel\mcite{ntk-metric}) that are potentially indicative of attack vulnerability as well.

\vspace{2pt}
{\bf Robustness, accuracy, and search efficiency --} It is revealed that the greater vulnerability incurred by \nas is possibly associated with the preference for models that converge fast at early training phases (\mie, higher trainability). It is however unclear whether this observation implies fundamental conflicts between the factors of robustness, accuracy, and search efficiency; if so, is it possible to find an optimal balance between them? We consider answering these questions critical for designing and operating \nas in practical settings.

\section{Related Work}
\label{sec:literature}

Next, we survey the literature relevant to this work.

\vspace{2pt}
{\bf Neural architecture search --} The existing \nas methods can be categorized along three dimensions: search space, search strategy, and performance measure.

The search space defines the possible set of candidate models. Early \nas methods focus on the chain-of-layer structure\mcite{rl-automl}, consisting of a sequence of layers. Motivated by that hand-crafted models often consist of repeated motifs, recent methods propose to search for such cell structures, including the connection topology and the corresponding operation on each connection\mcite{nasnet,enas,darts,snas,amoebanet}.  

The search strategy defines how to efficiently explore the pre-defined search space. Early \nas methods rely on either random search\mcite{random-perm} or Bayesian optimization\mcite{bayesian-nas}, which are often limited in terms of search efficiency and model complexity. More recent work mainly uses the approaches of reinforcement learning (RL)\mcite{rl-automl} or neural evolution\mcite{amoebanet,darts}. Empirically, neural evolution- and RL-based methods tend to perform comparably well\mcite{amoebanet}.

The performance measure evaluates the candidate models and guides the search process. Recently, one-shot \nas has emerged as a popular performance measure. It considers all candidate models as different sub-graphs of a super-net (\mie, the one-shot model) and shares weights between candidate models\mcite{enas,darts,snas}. The differentiable \nas methods considered in this paper belong to this category. Different one-shot methods differ in how the one-shot model is trained. For instance, {\darts}\mcite{darts} optimizes the one-shot model with  continuous relaxation of the search space.

\vspace{2pt}
{\bf ML Security --} With their wide use in security-sensitive domains, ML models are becoming the new targets for malicious manipulations\mcite{Biggio:2018:pr}. A variety of attack vectors have been exploited: adversarial evasion crafts adversarial inputs to force the target model to misbehave\mcite{goodfellow:fsgm,carlini-attack}; model poisoning modifies the target model's behavior (\meg, performance drop) via polluting its training data\mcite{model-reuse}; backdoor injection creates a trojan model such that any input embedded with a specific trigger is likely to be misclassified by the model\mcite{badnet,trojannn}; functionality stealing constructs a replicate model functionally similar to a victim model\mcite{knockoff-net,hardware-model-stealing}; membership inference breaches data privacy via inferring whether a given input is included in the model's training data based on the model's prediction\mcite{membership-attack}.

In response, another line of work strives to improve the resilience of ML models against such attacks. For instance, against adversarial evasion, existing defenses explore new training strategies (\meg, adversarial training)\mcite{pgd,Tramer:2018:iclr} and detection mechanisms\mcite{Meng:2017:ccs,Gehr:2018:sp}. Yet, such defenses often fail when facing even stronger attacks\mcite{Athalye:2018:icml,deepsec}, resulting in a constant arms race between the attackers and defenders.

\vspace{2pt}
Despite the intensive research on \nas and \ml security in parallel, the robustness of \nas-generated models to malicious manipulations is fairly under-explored\mcite{robust-nas}. Concurrent to this work, it is shown in\mcite{adversarial-robustness-arch} that \nas models tend to be more vulnerable to adversarial evasion, while our work differs in considering a variety of attacks beyond adversarial evasion, providing possible explanations for such vulnerability, and investigating potential mitigation.


\section{Conclusion}
\label{sec:conclusion}

This work represents a systematic study on the security risks incurred by \automl. From both empirical and analytical perspectives, we demonstrate that \nas-generated models tend to suffer greater vulnerability to various malicious manipulations, compared with their manually designed counterparts, which implies the existence of fundamental drawbacks in the design of existing \nas methods. We identify high loss smoothness and low gradient variance, stemming from the preference of \nas for models with higher trainability, as possible causes for such phenomena. Our findings raise concerns about the current practice of \nas in security-sensitive domains. Further, we discuss potential remedies to mitigate such limitations, which sheds light on designing and operating \nas in a more robust and principled manner.

\newpage 



\bibliographystyle{plain}
\bibliography{bibs/aml,bibs/debugging,bibs/general,bibs/optimization,bibs/main,bibs/automl}

\begin{thebibliography}{10}

\bibitem{andrychowicz:nips:2016}
Marcin Andrychowicz, Misha Denil, Sergio G\'{o}mez, Matthew~W Hoffman, David
  Pfau, Tom Schaul, Brendan Shillingford, and Nando de~Freitas.
\newblock {Learning to Learn by Gradient Descent by Gradient Descent}.
\newblock In {\em Proceedings of Advances in Neural Information Processing
  Systems (NeurIPS)}, 2016.

\bibitem{Athalye:2018:icml}
Anish {Athalye}, Nicholas {Carlini}, and David {Wagner}.
\newblock {Obfuscated Gradients Give a False Sense of Security: Circumventing
  Defenses to Adversarial Examples}.
\newblock In {\em Proceedings of IEEE Conference on Machine Learning (ICML)},
  2018.

\bibitem{rl-automl}
Bowen {Baker}, Otkrist {Gupta}, Nikhil {Naik}, and Ramesh {Raskar}.
\newblock {Designing Neural Network Architectures using Reinforcement
  Learning}.
\newblock In {\em Proceedings of International Conference on Learning
  Representations (ICLR)}, 2017.

\bibitem{bayesian-nas}
J.~Bergstra, D.~Yamins, and D.~D. Cox.
\newblock {Making a Science of Model Search: Hyperparameter Optimization in
  Hundreds of Dimensions for Vision Architectures}.
\newblock In {\em Proceedings of IEEE Conference on Machine Learning (ICML)},
  2013.

\bibitem{Biggio:2012:icml}
Battista {Biggio}, Blaine {Nelson}, and Pavel {Laskov}.
\newblock {Poisoning Attacks against Support Vector Machines}.
\newblock In {\em Proceedings of IEEE Conference on Machine Learning (ICML)},
  2012.

\bibitem{Biggio:2018:pr}
Battista Biggio and Fabio Roli.
\newblock {Wild Patterns: Ten Years after The Rise of Adversarial Machine
  Learning}.
\newblock {\em Pattern Recognition}, 84:317--331, 2018.

\bibitem{path-level}
Han {Cai}, Jiacheng {Yang}, Weinan {Zhang}, Song {Han}, and Yong {Yu}.
\newblock {Path-Level Network Transformation for Efficient Architecture
  Search}.
\newblock In {\em Proceedings of IEEE Conference on Machine Learning (ICML)},
  2018.

\bibitem{carlini-attack}
Nicholas Carlini and David~A. Wagner.
\newblock {Towards Evaluating the Robustness of Neural Networks}.
\newblock In {\em Proceedings of IEEE Symposium on Security and Privacy
  (S\&P)}, 2017.

\bibitem{hopskipjump}
Jianbo {Chen}, Michael~I. {Jordan}, and Martin~J. {Wainwright}.
\newblock {HopSkipJumpAttack: A Query-Efficient Decision-Based Attack}.
\newblock In {\em Proceedings of IEEE Symposium on Security and Privacy
  (S\&P)}, 2020.

\bibitem{ntk-metric}
Wuyang Chen, Xinyu Gong, and Zhangyang Wang.
\newblock {Neural Architecture Search on ImageNet in Four {\{}GPU{\}} Hours: A
  Theoretically Inspired Perspective}.
\newblock In {\em Proceedings of International Conference on Learning
  Representations (ICLR)}, 2021.

\bibitem{drnas}
Xiangning {Chen}, Ruochen {Wang}, Minhao {Cheng}, Xiaocheng {Tang}, and Cho-Jui
  {Hsieh}.
\newblock {DrNAS: Dirichlet Neural Architecture Search}.
\newblock In {\em Proceedings of International Conference on Learning
  Representations (ICLR)}, 2021.

\bibitem{pdarts}
Xin {Chen}, Lingxi {Xie}, Jun {Wu}, and Qi~{Tian}.
\newblock {Progressive Differentiable Architecture Search: Bridging the Depth
  Gap between Search and Evaluation}.
\newblock In {\em Proceedings of IEEE International Conference on Computer
  Vision (ICCV)}, 2019.

\bibitem{label-only-membership}
Christopher~A. {Choquette-Choo}, Florian {Tramer}, Nicholas {Carlini}, and
  Nicolas {Papernot}.
\newblock {Label-Only Membership Inference Attacks}.
\newblock {\em ArXiv e-prints}, 2020.

\bibitem{darts-}
Xiangxiang {Chu}, Xiaoxing {Wang}, Bo~{Zhang}, Shun {Lu}, Xiaolin {Wei}, and
  Junchi {Yan}.
\newblock {DARTS-: Robustly Stepping out of Performance Collapse Without
  Indicators}.
\newblock In {\em Proceedings of International Conference on Learning
  Representations (ICLR)}, 2021.

\bibitem{imgnet}
J.~Deng, W.~Dong, R.~Socher, L.~Li, Kai Li, and Li~Fei-Fei.
\newblock {ImageNet: A Large-scale Hierarchical Image Database}.
\newblock In {\em Proceedings of IEEE Conference on Computer Vision and Pattern
  Recognition (CVPR)}, 2009.

\bibitem{adversarial-robustness-arch}
Chaitanya {Devaguptapu}, Devansh {Agarwal}, Gaurav {Mittal}, Pulkit {Gopalani},
  and Vineeth~N {Balasubramanian}.
\newblock {On Adversarial Robustness: A Neural Architecture Search
  perspective}.
\newblock In {\em RobustML Workshop of International Conference on Learning
  Representations}, 2021.

\bibitem{nas-bench}
Xuanyi Dong and Yi~Yang.
\newblock {NAS-Bench-201: Extending the Scope of Reproducible Neural
  Architecture Search}.
\newblock In {\em Proceedings of International Conference on Learning
  Representations (ICLR)}, 2020.

\bibitem{automl-survey}
Thomas {Elsken}, Jan {Hendrik Metzen}, and Frank {Hutter}.
\newblock {Neural Architecture Search: A Survey}.
\newblock {\em Journal of Machine Learning Research}, (20):1--21, 2019.

\bibitem{Gehr:2018:sp}
T.~Gehr, M.~Mirman, D.~Drachsler-Cohen, P.~Tsankov, S.~Chaudhuri, and
  M.~Vechev.
\newblock {AI2: Safety and Robustness Certification of Neural Networks with
  Abstract Interpretation}.
\newblock In {\em Proceedings of IEEE Symposium on Security and Privacy
  (S\&P)}, 2018.

\bibitem{gradient-deviation}
Saeed {Ghadimi} and Guanghui {Lan}.
\newblock {Stochastic First- and Zeroth-order Methods for Nonconvex Stochastic
  Programming}.
\newblock {\em SIAM Journal on Optimization}, 23(4):2341--2368, 2013.

\bibitem{goodfellow:fsgm}
Ian Goodfellow, Jonathon Shlens, and Christian Szegedy.
\newblock {Explaining and Harnessing Adversarial Examples}.
\newblock In {\em Proceedings of International Conference on Learning
  Representations (ICLR)}, 2015.

\bibitem{loss-landscape}
Ian~J. {Goodfellow}, Oriol {Vinyals}, and Andrew~M. {Saxe}.
\newblock {Qualitatively Characterizing Neural Network Optimization Problems}.
\newblock In {\em Proceedings of Advances in Neural Information Processing
  Systems (NeurIPS)}, 2015.

\bibitem{badnet}
Tianyu {Gu}, Brendan {Dolan-Gavitt}, and Siddharth {Garg}.
\newblock {BadNets: Identifying Vulnerabilities in the Machine Learning Model
  Supply Chain}.
\newblock {\em ArXiv e-prints}, 2017.

\bibitem{robust-nas}
Minghao {Guo}, Yuzhe {Yang}, Rui {Xu}, Ziwei {Liu}, and Dahua {Lin}.
\newblock {When NAS Meets Robustness: In Search of Robust Architectures against
  Adversarial Attacks}.
\newblock In {\em Proceedings of IEEE Conference on Computer Vision and Pattern
  Recognition (CVPR)}, 2019.

\bibitem{kaiming-init}
Kaiming {He}, Xiangyu {Zhang}, Shaoqing {Ren}, and Jian {Sun}.
\newblock {Delving Deep into Rectifiers: Surpassing Human-Level Performance on
  ImageNet Classification}.
\newblock In {\em Proceedings of IEEE International Conference on Computer
  Vision (ICCV)}, 2015.

\bibitem{resnet}
Kaiming He, Xiangyu Zhang, Shaoqing Ren, and Jian Sun.
\newblock {Deep Residual Learning for Image Recognition}.
\newblock In {\em Proceedings of IEEE Conference on Computer Vision and Pattern
  Recognition (CVPR)}, 2016.

\bibitem{hardware-model-stealing}
Sanghyun {Hong}, Michael {Davinroy}, Yi{\u{g}}itcan {Kaya}, Dana
  {Dachman-Soled}, and Tudor {Dumitra{\c{s}}}.
\newblock {How to 0wn NAS in Your Spare Time}.
\newblock In {\em Proceedings of International Conference on Learning
  Representations (ICLR)}, 2020.

\bibitem{densenet}
Gao {Huang}, Zhuang {Liu}, Laurens {van der Maaten}, and Kilian~Q.
  {Weinberger}.
\newblock {Densely Connected Convolutional Networks}.
\newblock In {\em Proceedings of IEEE Conference on Computer Vision and Pattern
  Recognition (CVPR)}, 2017.

\bibitem{Ilyas:2018:icml}
Andrew Ilyas, Logan Engstrom, Anish Athalye, and Jessy Lin.
\newblock {Black-box Adversarial Attacks with Limited Queries and Information}.
\newblock In {\em Proceedings of IEEE Conference on Machine Learning (ICML)},
  2018.

\bibitem{model-reuse}
Yujie {Ji}, Xinyang {Zhang}, Shouling {Ji}, Xiapu {Luo}, and Ting {Wang}.
\newblock {Model-Reuse Attacks on Deep Learning Systems}.
\newblock In {\em Proceedings of ACM SAC Conference on Computer and
  Communications (CCS)}, 2018.

\bibitem{random-perm}
Rafal Jozefowicz, Wojciech Zaremba, and Ilya Sutskever.
\newblock {An Empirical Exploration of Recurrent Network Architectures}.
\newblock In {\em Proceedings of IEEE Conference on Machine Learning (ICML)},
  2015.

\bibitem{bit-net}
Alexander {Kolesnikov}, Lucas {Beyer}, Xiaohua {Zhai}, Joan {Puigcerver},
  Jessica {Yung}, Sylvain {Gelly}, and Neil {Houlsby}.
\newblock {Big Transfer (BiT): General Visual Representation Learning}.
\newblock In {\em Proceedings of European Conference on Computer Vision
  (ECCV)}, 2020.

\bibitem{cifar}
Alex Krizhevsky and Geoffrey Hinton.
\newblock {Learning Multiple Layers of Features from Tiny Images}.
\newblock {\em Technical report, University of Toronto}, 2009.

\bibitem{wide-dnn}
Jaehoon {Lee}, Lechao {Xiao}, Samuel~S. {Schoenholz}, Yasaman {Bahri}, Roman
  {Novak}, Jascha {Sohl-Dickstein}, and Jeffrey {Pennington}.
\newblock {Wide Neural Networks of Any Depth Evolve as Linear Models under
  Gradient Descent}.
\newblock In {\em Proceedings of Advances in Neural Information Processing
  Systems (NeurIPS)}, 2020.

\bibitem{sgas}
Guohao {Li}, Guocheng {Qian}, Itzel~C. {Delgadillo}, Matthias {M{\"u}ller}, Ali
  {Thabet}, and Bernard {Ghanem}.
\newblock {SGAS: Sequential Greedy Architecture Search}.
\newblock In {\em Proceedings of International Conference on Learning
  Representations (ICLR)}, 2020.

\bibitem{loss-vis}
Hao {Li}, Zheng {Xu}, Gavin {Taylor}, Christoph {Studer}, and Tom {Goldstein}.
\newblock {Visualizing the Loss Landscape of Neural Nets}.
\newblock In {\em Proceedings of Advances in Neural Information Processing
  Systems (NeurIPS)}, 2018.

\bibitem{learn-to-optimize}
Ke~{Li} and Jitendra {Malik}.
\newblock {Learning to Optimize}.
\newblock In {\em Proceedings of International Conference on Learning
  Representations (ICLR)}, 2017.

\bibitem{deepsec}
X.~Ling, S.~Ji, J.~Zou, J.~Wang, C.~Wu, B.~Li, and T.~Wang.
\newblock {DEEPSEC: A Uniform Platform for Security Analysis of Deep Learning
  Model}.
\newblock In {\em Proceedings of IEEE Symposium on Security and Privacy
  (S\&P)}, 2019.

\bibitem{darts}
Hanxiao {Liu}, Karen {Simonyan}, and Yiming {Yang}.
\newblock {DARTS: Differentiable Architecture Search}.
\newblock In {\em Proceedings of International Conference on Learning
  Representations (ICLR)}, 2019.

\bibitem{trojannn}
Yingqi Liu, Shiqing Ma, Yousra Aafer, Wen-Chuan Lee, Juan Zhai, Weihang Wang,
  and Xiangyu Zhang.
\newblock {Trojaning Attack on Neural Networks}.
\newblock In {\em Proceedings of Network and Distributed System Security
  Symposium (NDSS)}, 2018.

\bibitem{Madry:2017:iclr}
A.~{Madry}, A.~{Makelov}, L.~{Schmidt}, D.~{Tsipras}, and A.~{Vladu}.
\newblock {Towards Deep Learning Models Resistant to Adversarial Attacks}.
\newblock In {\em Proceedings of the International Conference on Learning
  Representations (ICLR)}, 2017.

\bibitem{pgd}
Aleksander Madry, Aleksandar Makelov, Ludwig Schmidt, Dimitris Tsipras, and
  Adrian Vladu.
\newblock {Towards Deep Learning Models Resistant to Adversarial Attacks}.
\newblock In {\em Proceedings of International Conference on Learning
  Representations (ICLR)}, 2018.

\bibitem{Meng:2017:ccs}
Dongyu Meng and Hao Chen.
\newblock Magnet: A two-pronged defense against adversarial examples.
\newblock In {\em Proceedings of the ACM SIGSAC Conference on Computer and
  Communications Security (CCS)}, 2017.

\bibitem{knockoff-net}
Tribhuvanesh {Orekondy}, Bernt {Schiele}, and Mario {Fritz}.
\newblock {Knockoff Nets: Stealing Functionality of Black-Box Models}.
\newblock In {\em Proceedings of IEEE Conference on Computer Vision and Pattern
  Recognition (CVPR)}, 2018.

\bibitem{evil-twin}
Ren {Pang}, Hua {Shen}, Xinyang {Zhang}, Shouling {Ji}, Yevgeniy {Vorobeychik},
  Xiapu {Luo}, Alex {Liu}, and Ting {Wang}.
\newblock {A Tale of Evil Twins: Adversarial Inputs versus Poisoned Models}.
\newblock In {\em Proceedings of ACM SAC Conference on Computer and
  Communications (CCS)}, 2020.

\bibitem{enas}
Hieu {Pham}, Melody~Y. {Guan}, Barret {Zoph}, Quoc~V. {Le}, and Jeff {Dean}.
\newblock {Efficient Neural Architecture Search via Parameter Sharing}.
\newblock In {\em Proceedings of IEEE Conference on Machine Learning (ICML)},
  2018.

\bibitem{amoebanet}
Esteban {Real}, Alok {Aggarwal}, Yanping {Huang}, and Quoc~V {Le}.
\newblock {Regularized Evolution for Image Classifier Architecture Search}.
\newblock In {\em Proceedings of AAAI Conference on Artificial Intelligence
  (AAAI)}, 2019.

\bibitem{weight-normalization}
Tim {Salimans} and Diederik~P. {Kingma}.
\newblock {Weight normalization: A simple reparameterization to accelerate
  training of deep neural networks}.
\newblock In {\em Proceedings of Advances in Neural Information Processing
  Systems (NeurIPS)}, 2016.

\bibitem{adv-train-free}
Ali {Shafahi}, Mahyar {Najibi}, Amin {Ghiasi}, Zheng {Xu}, John {Dickerson},
  Christoph {Studer}, Larry~S. {Davis}, Gavin {Taylor}, and Tom {Goldstein}.
\newblock {Adversarial Training for Free!}
\newblock In {\em Proceedings of Advances in Neural Information Processing
  Systems (NeurIPS)}, 2019.

\bibitem{membership-attack}
Reza {Shokri}, Marco {Stronati}, Congzheng {Song}, and Vitaly {Shmatikov}.
\newblock {Membership Inference Attacks against Machine Learning Models}.
\newblock In {\em Proceedings of IEEE Symposium on Security and Privacy
  (S\&P)}, 2017.

\bibitem{shallow-wide-cell}
Yao {Shu}, Wei {Wang}, and Shaofeng {Cai}.
\newblock {Understanding Architectures Learnt by Cell-based Neural Architecture
  Search}.
\newblock In {\em Proceedings of International Conference on Learning
  Representations (ICLR)}, 2020.

\bibitem{vgg}
Karen Simonyan and Andrew Zisserman.
\newblock {Very Deep Convolutional Networks for Large-Scale Image Recognition}.
\newblock In {\em Proceedings of International Conference on Learning
  Representations (ICLR)}, 2014.

\bibitem{Tramer:2018:iclr}
F.~{Tram{\`e}r}, A.~{Kurakin}, N.~{Papernot}, I.~{Goodfellow}, D.~{Boneh}, and
  P.~{McDaniel}.
\newblock {Ensemble Adversarial Training: Attacks and Defenses}.
\newblock In {\em Proceedings of International Conference on Learning
  Representations (ICLR)}, 2018.

\bibitem{model-stealing}
Florian Tram\`{e}r, Fan Zhang, Ari Juels, Michael~K. Reiter, and Thomas
  Ristenpart.
\newblock {Stealing Machine Learning Models via Prediction APIs}.
\newblock In {\em Proceedings of USENIX Security Symposium (SEC)}, 2016.

\bibitem{rethinking-nas}
Ruochen Wang, Minhao Cheng, Xiangning Chen, Xiaocheng Tang, and Cho-Jui Hsieh.
\newblock {Rethinking Architecture Selection in Differentiable {NAS}}.
\newblock In {\em Proceedings of International Conference on Learning
  Representations (ICLR)}, 2021.

\bibitem{skip-connection}
Dongxian Wu, Yisen Wang, Shu-Tao Xia, James Bailey, and Xingjun Ma.
\newblock {Skip Connections Matter: On the Transferability of Adversarial
  Examples Generated with ResNets}.
\newblock In {\em Proceedings of International Conference on Learning
  Representations (ICLR)}, 2020.

\bibitem{resnext}
Saining {Xie}, Ross {Girshick}, Piotr {Doll{\'a}r}, Zhuowen {Tu}, and Kaiming
  {He}.
\newblock {Aggregated Residual Transformations for Deep Neural Networks}.
\newblock In {\em Proceedings of IEEE Conference on Computer Vision and Pattern
  Recognition (CVPR)}, 2017.

\bibitem{snas}
Sirui {Xie}, Hehui {Zheng}, Chunxiao {Liu}, and Liang {Lin}.
\newblock {SNAS: Stochastic Neural Architecture Search}.
\newblock In {\em Proceedings of International Conference on Learning
  Representations (ICLR)}, 2019.

\bibitem{pcdarts}
Yuhui {Xu}, Lingxi {Xie}, Xiaopeng {Zhang}, Xin {Chen}, Guo-Jun {Qi},
  Qi~{Tian}, and Hongkai {Xiong}.
\newblock {PC-DARTS: Partial Channel Connections for Memory-Efficient
  Architecture Search}.
\newblock In {\em Proceedings of International Conference on Learning
  Representations (ICLR)}, 2020.

\bibitem{dla}
Fisher {Yu}, Dequan {Wang}, Evan {Shelhamer}, and Trevor {Darrell}.
\newblock {Deep Layer Aggregation}.
\newblock In {\em Proceedings of IEEE Conference on Computer Vision and Pattern
  Recognition (CVPR)}, 2018.

\bibitem{wideresnet}
Sergey {Zagoruyko} and Nikos {Komodakis}.
\newblock {Wide Residual Networks}.
\newblock In {\em Proceedings of British Machine Vision Conference (BMVC)},
  2016.

\bibitem{darts-r}
Arber {Zela}, Thomas {Elsken}, Tonmoy {Saikia}, Yassine {Marrakchi}, Thomas
  {Brox}, and Frank {Hutter}.
\newblock {Understanding and Robustifying Differentiable Architecture Search}.
\newblock In {\em Proceedings of International Conference on Learning
  Representations (ICLR)}, 2020.

\bibitem{rl-nas-c}
Zhao {Zhong}, Junjie {Yan}, Wei {Wu}, Jing {Shao}, and Cheng-Lin {Liu}.
\newblock {Practical Block-wise Neural Network Architecture Generation}.
\newblock In {\em Proceedings of IEEE Conference on Computer Vision and Pattern
  Recognition (CVPR)}, 2018.

\bibitem{nasnet}
Barret {Zoph}, Vijay {Vasudevan}, Jonathon {Shlens}, and Quoc~V. {Le}.
\newblock {Learning Transferable Architectures for Scalable Image Recognition}.
\newblock In {\em Proceedings of IEEE Conference on Computer Vision and Pattern
  Recognition (CVPR)}, 2018.

\end{thebibliography}

\appendix

\section{Notations}

\begin{table}[!ht]{\footnotesize
\centering
\renewcommand{\arraystretch}{1.2}
			\begin{tabular}{c | l }
				Notation & Definition\\
				\hline
				\hline
				$f$, $x$ & model, input  \\
				$\ell(\cdot, \cdot)$  & loss function w.r.t. a single input\\
				$\gL(\cdot)$ & overall loss function w.r.t. a dataset\\
				$\ssup{o}{(i,j)}$  & operation on edge $(i,j)$\\
				$\sboth{\alpha}{o}{(i,j)}$ & weight of operation $o$ on edge  $(i,j)$\\
				$\alpha$, $\theta$ & architecture and model parameters\\
				$\| \cdot \|$ & vector or matrix norm\\
				$\ssub{\gD}{\text{trn}}$, $\ssub{\gD}{\text{pos}}$, $\ssub{\gD}{\text{tst}}$ & training, poisoning, testing set\\
				\hline
			\end{tabular}
    \caption{Symbols and notations. \label{tab:notations}}}
\end{table}

\section{Proofs}

\label{sec:proof}

\begin{proof}(Theorem\mref{the:smoothness}) Without loss of generality, we assume the loss function $\gL$ is computed over a single input-label pair: 
\begin{equation}
\gL(x;\theta) = \ell(f(x;\theta), y)
\end{equation}

We split the model $f$ into its first layer and the remaining layers (with parameters $\bar{\theta}$). Typically, the first layer (without the non-linear activation) can be modeled as a linear function $Ax + b$ (\meg, fully-connected or convolutional layer). We thus rewrite $\gL$ as a composite function:
$\gL(x;\theta) = \bar{\gL}( Ax + b; \bar{\theta})$, where $\bar{\gL}$ (parameterized by $\bar{\theta}$) is $\gL$ excluding $f$'s first layer.

According to the assumption that $\gL(x;\theta)$ has $L$-Lipschitz continuous gradient with respect to $\theta$, with $\bar{\theta}$ and $A$ fixed,
\begin{equation}
\|\nabla_b \bar{\gL}(Ax+b) \|= \| \nabla \bar{\gL} \| \leq L
\end{equation}
Thus, the gradient of $\gL$ with respect to $x$ is also bounded:
\begin{equation}
\|\nabla_x \bar{\gL}(Ax+b)\|=\|\ssup{A}{\intercal} \nabla \bar{\gL} \| \leq  \|\ssup{A}{\intercal}\|\|\nabla \bar{\gL}\| \leq L\|\ssup{A}{\intercal}\|
\end{equation}

With weight normalization, $\forall i$ $\sum_j {A_{ij}}=0$, $\sum_j {A_{ij}^2}=1$. Applying the Chebyshev's inequality, we bound $\|\ssup{A}{\intercal}\|_1$ as:
\begin{equation}
\hspace{-2pt}
\|\ssup{A}{\intercal}\|_1=\|A\|_\infty=\max_{1\leq i\leq n}\sum_j {|A_{ij}|}\leq \max_{1\leq i\leq n}\sqrt{\frac{\sum_j {A_{ij}^2}}{n}}=\frac{1}{\sqrt{n}}
\end{equation}
where $n$ is the number of rows of $A$ (\mie, the input dimensionality). Putting everything together,
\begin{equation}
\|\nabla_x \gL(x;\theta)\|_1 \leq \frac{L}{\sqrt{n}}
\end{equation}

Therefore, $\gL(x;\theta)$ has $\frac{L}{\sqrt{n}}$-Lipschitz continuous gradient with respect to input $x$.

\end{proof}

\section{Parameter Setting}
\label{sec:appb}


Table\mref{tab:setting} summarizes the default parameter setting in \msec{sec:measure}.

\begin{table*}[!ht]{\footnotesize
\renewcommand{\arraystretch}{1.2}
\setlength{\tabcolsep}{3pt}
\centering
\begin{tabular}{r|r|l}
  Type & Parameter & Setting \\
  \hline
  \hline
  \multirow{6}{*}{Training} & Optimizer & SGD \\
  & Initial learning rate & 0.025 \\
  & LR scheduler & Cosine annealing  \\
  & Gradient clipping threshold & 5.0 \\
  & Training epochs & 600 \\
  & Batch size & 96\\
  \hline
  \hline
  \multirow{5}{*}{Adversarial Evasion} 
  & Perturbation threshold & $\epsilon=8/255$ \\
  & Learning rate & $\alpha=2/255$ \\
  & Maximum iterations (M) & 3 \\
  & Maximum iterations (L) & 7 \\
  & Number of random restarts & 5\\
  \hline
  Model Poisoning & Training epochs & 50 \\
  \hline
            \multirow{7}{*}{Backdoor Injection}     & Pre-processing layer          & Penultimate     \\
                                      & Number of target neurons             & 2                       \\
                                      & Pre-processing optimizer      & PGD                     \\
                                      & Pre-processing learning rate             & 0.015                   \\
                                      & Pre-processing iterations          & 20                      \\
                                      & Trigger size  &3$\times$3 \\
                                      & Trigger transparency & 0.7\\
            \hline
 \multirow{3}{*}{Functionality Stealing}   &  Sampling strategy & Adaptive\\
 & Training epochs & 50 \\
  & Reward type & All\\
  
  \hline 
     \multirow{5}{*}{Membership Inference}  
     & $\ell_p$-norm  & 2\\
     &  Maximum iterations & 50\\
 & Maximum evaluation & 2,500 \\
  & Initial evaluation & 100\\
  & Initial size & 100\\
  \hline
  \hline
  \multirow{3}{*}{Adversarial Training} 
  & Perturbation threshold & $\epsilon=8/255$ \\
  & Learning rate & $\alpha=2/255$ \\
  & Perturbation iterations & 7 \\
    
\end{tabular}
\caption{Default parameter setting used in \msec{sec:measure} (M - most likely case; L - least likely case).
\label{tab:setting}}}
\end{table*}

\section{Cell Structures of \nas Models}

\begin{figure*}[!ht]
    \centering
    \includegraphics[width=172mm]{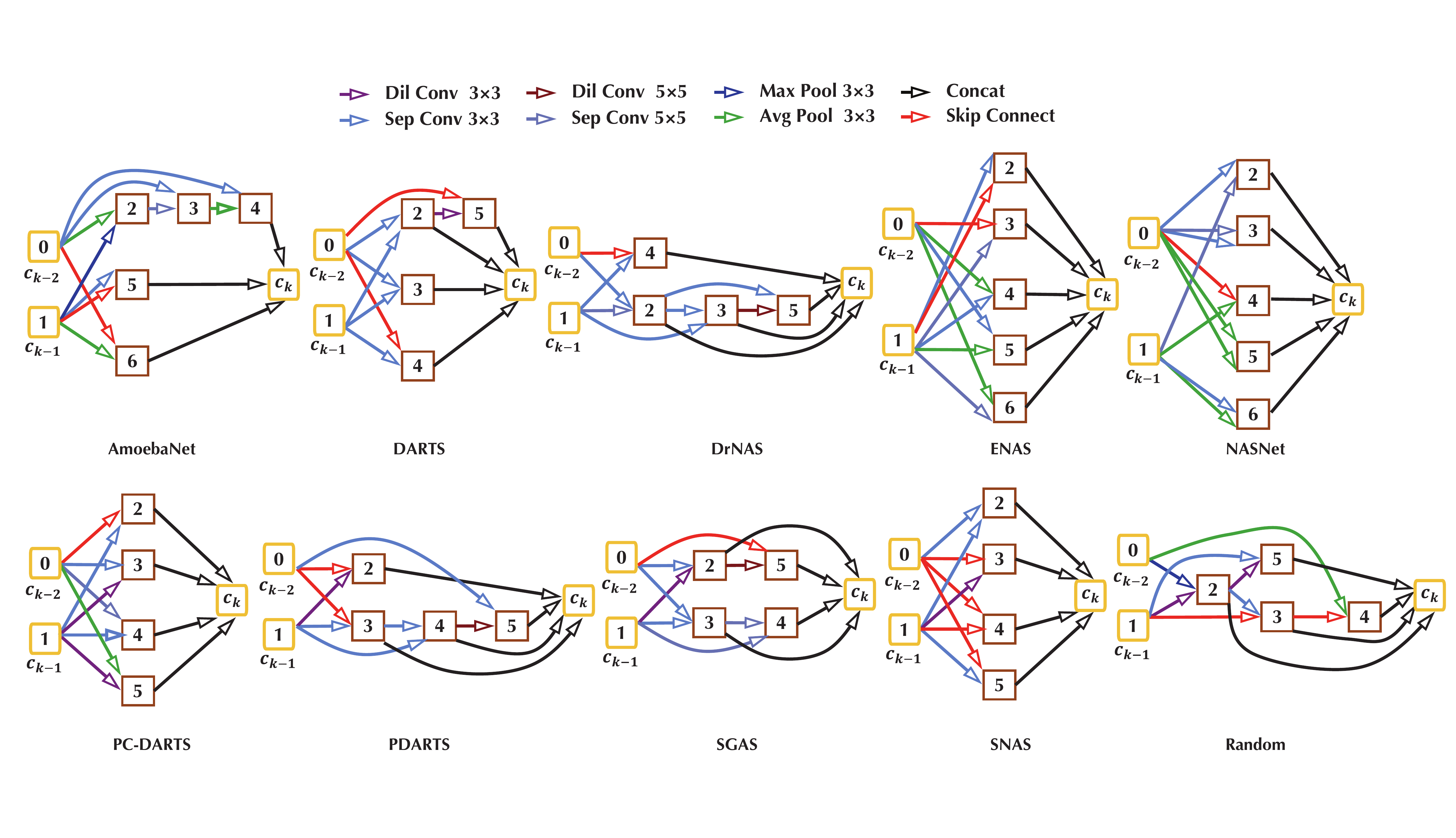}
    \caption{Cell structures of NAS-generated models. \label{fig:arch_nas}}
\end{figure*}

Figure\mref{fig:arch_nas} depicts the cell structures generated by the \nas methods on \cifar.

\end{document}